\documentclass{article}

\usepackage{arxiv}

\usepackage{amsmath}
\usepackage{graphicx}
\usepackage[round]{natbib}
\usepackage{url} 

\usepackage{amsmath}
\usepackage{amsfonts}
\usepackage[colorlinks=true, allcolors=blue]{hyperref}
\usepackage{dsfont}
\usepackage{multirow}
\usepackage{array}
\newcolumntype{C}[1]{>{\centering\arraybackslash}m{#1}}
\newcolumntype{L}[1]{>{\arraybackslash}m{#1}}
\usepackage[export]{adjustbox}

\usepackage{makecell}
\usepackage{comment}
\usepackage{multicol}
\usepackage{diagbox}

\usepackage{tikz}
\usetikzlibrary{bayesnet}

\usetikzlibrary{positioning}
\usetikzlibrary{matrix,positioning,arrows.meta,arrows,trees}
\tikzset{
mymat/.style={
  matrix of math nodes,
  text height=2.5ex,
  text depth=0.75ex,
  text width=3.25ex,
  align=center,
  column sep=-\pgflinewidth
  },
mymats/.style={
  mymat,
  nodes={draw,fill=#1}
  }  
}
\usepackage{mathtools}

\DeclarePairedDelimiter\ceil{\lceil}{\rceil}
\usepackage{subcaption}

\usepackage{twoopt}
\usetikzlibrary{fit}
\newcommand\tikzmark[1]{%
  \tikz[remember picture,overlay]\node[inner xsep=0pt] (#1) {};}
\newcommandtwoopt\Textbox[5][2.5cm][2cm]{%
\begin{tikzpicture}[remember picture,overlay]
  \coordinate (aux) at ([xshift=#1]#4);
  \node[inner ysep=5pt,yshift=0.ex,draw=red,thick,
    fit=(#3) (aux),baseline] 
    (box) {};
  \node[text width=#2,anchor=north east,
    font=\sffamily\footnotesize,align=right] 
    at (box.north east) {#5};
\end{tikzpicture}%
}

\usepackage{amsthm}
\newtheorem{theorem}{Theorem}
\newtheorem{proposition}{Proposition}
\newtheorem{lemma}{Lemma}
\newtheorem{definition}{Definition}

\newdimen\nodeDist
\usepackage{algorithm}
\usepackage{algpseudocode}
\usepackage[table]{xcolor}
\title{Efficient distributional regression trees learning algorithms for calibrated non-parametric probabilistic forecasts}

\author{ \href{https://orcid.org/0000-0003-3636-3770}{\includegraphics[scale=0.06]{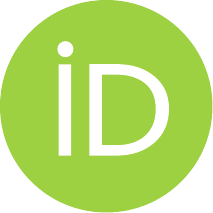}\hspace{1mm}Quentin Duchemin} \\
	Swiss Data Science Center \\ EPFL \& ETH Zürich\\ Switzerland\\
	\texttt{quentin.duchemin@epfl.ch} \\
	\And
	\href{https://orcid.org/0000-0002-7629-7571}{\includegraphics[scale=0.06]{orcid.pdf}\hspace{1mm}Guillaume Obozinski} \\
	Swiss Data Science Center \\ EPFL \& ETH Zürich\\ Switzerland \\
	\texttt{guillaume.obozinski@epfl.ch} \\
}

\begin{document}

\def\spacingset#1{\renewcommand{\baselinestretch}%
{#1}\small\normalsize} \spacingset{1}


\maketitle

\begin{abstract}
The perspective of developing trustworthy AI for critical applications in science and engineering requires machine learning techniques that are capable of estimating their \emph{own uncertainty}. 
In the context of regression, instead of estimating a conditional mean, this can be achieved by producing a predictive interval for the output, or to even learn a model of the conditional probability $p(y|x)$ of an output $y$ given input features $x$. 
While this can be done under parametric assumptions with, e.g.\ generalized linear model, these are typically too strong, and non-parametric models offer flexible alternatives. In particular, for scalar outputs, learning directly a model of the \emph{conditional cumulative distribution function} of $y$ given $x$ can lead to more precise probabilistic estimates, and the use of proper scoring rules such as the \emph{weighted interval score} (WIS) and the \emph{continuous ranked probability score} (CRPS) lead to better coverage and calibration properties.

This paper introduces novel algorithms for learning probabilistic regression trees for the WIS or CRPS loss functions.
These algorithms are made computationally efficient thanks to an appropriate use of known data structures - namely \emph{min-max heaps, weight-balanced binary trees} and \emph{Fenwick trees}.

Through numerical experiments, we demonstrate that the performance of our methods is competitive with alternative approaches. 
\end{abstract}

\noindent%

\vfill

\newpage
\spacingset{1.75} 

\section{Introduction}

Unlike standard regression methods which target the conditional mean of the response to explanatory variables, distributional regression aims at estimating the entire conditional distribution of the outcome. Distributional regression is becoming more and more popular since understanding variability and uncertainty is crucial in fields like econometrics, climatology, and biostatistics. \cite{rigby2005} laid foundational work with their generalized additive models for location, scale, and shape, in which a parametric distribution is assumed for the response variable but the parameters of this distribution can vary according to explanatory variables using linear, nonlinear or smooth functions. Another line of work relies on density estimation techniques to learn the conditional distributions with methods ranging from mixture models \citep{grun2007fitting} to Bayesian ones \citep{dunson2007bayesian}. More recently, the deep learning area gave birth to generative models achieving remarkable results, especially in the field of imaging \citep{rombach2022high}. These generative models aim at learning to sample from the training data distribution via a transformation of a
simple distribution. One can mention normalizing flows \citep{kobyzev2020normalizing} or diffusion models \citep{ho2020denoising}. These methods have been widely applied to images and text, but recent works \citep{shen2024} adapt these methods to classical regression. We refer to a recent review of distributional regression approaches in \cite{kneib2023}.

While distributional regression provides a comprehensive view of the response distribution, there are cases where practitioners are interested in specific aspects of this distribution, rather than the entire probability landscape. For instance, understanding the behavior of a response variable at its extremes or at specific percentiles can be crucial for applications such as risk assessment or resource allocation. This is where quantile regression \citep{koenker1978regression,regression2017handbook} becomes a powerful complementary tool. Quantile regression has become a standard technique and is usually based  on the minimization of the so-called pinball loss. The pinball loss $\ell_{\tau}$ for the quantile of level $\tau \in (0,1)$ is defined by
\[\ell_{\tau}(\xi) = (\tau- \mathds 1_{\xi <0})\xi,\quad \xi \in \mathds R\]and is such that
\[q_{\tau}(x) \in {\arg \min}_{q \in \mathds R} \mathds E_P[ \ell_{\tau} (Y -q) \mid X=x],\]
where $q_{\tau}(x)$ is the quantile of level $\tau$ of the distribution of $Y$ conditional on $X=x$. While the pinball loss is not differentiable at zero, optimization can still be performed using subgradient methods or by employing a smoothed approximation, such as the Huber loss for the $\ell^1$ loss. This approach makes it possible to use of standard regression models to estimate specific conditional quantiles of interest. Notable examples include the use of neural networks \citep{CANNON20111277}, kernel methods \citep{Takeuchi20061231}, Gaussian processes \citep{Abeywardana20151686} or splines \citep{thompson2010bayesian}. We refer to the survey paper \cite{torossian2020} for a more exhaustive list of quantile regression methods.

Simultaneous multiple quantile regression provides a natural bridge between full distributional regression and single quantile regression. Estimating a set of quantiles rather than a single one can be important in various applications such as decision making \citep{Hamilton1999}, personalized health \citep{kulasekera2022quantiles} or finance \citep{Engle2004} where professionals often need to understand the potential losses or gains at various quantiles to assess the risk of different investments or portfolios. A straightforward approach to multiple quantile regression involves fitting separate models for each quantile of interest. However, this method can be suboptimal in practice, as it fails to leverage the shared information between the models, potentially leading to poorer performance. Additionally, it may result in the undesirable phenomenon of quantile crossing, where estimated quantiles are inconsistent, namely when higher level quantiles are lower than lower level quantiles. To address these issues, regularized approaches \citep{ZOU20085296} or constrained methods \citep{liu2009stepwise} have been proposed to couple different quantile models. To improve computational efficiency, a promising strategy is to optimize the sum of pinball losses across multiple quantiles in a single model \citep{juban2016multiple}, rather than fitting separate models for each quantile. The sum of pinball losses is known as the weighted interval score (WIS) in the probabilistic forecasting community \citep{WIS}. Given $M$ quantile levels $0<\tau_1<\dots <\tau_M<1/2$, the WIS loss is defined by:
\[\ell_{WIS(\tau_{1:M})}(\{q_{\tau_m}\}_{m=1}^{2M+1}, y)=\frac{2}{2M+1} \sum_{m=1}^{2M+1} \ell_{\tau_m}(y-q_{\tau_m}),\]
where $\tau_{M+1}=1/2$ and where $\tau_{M+2}=1-\tau_{M}$, ..., $\tau_{2M+1}=1-\tau_{1}$. The WIS is primarily a loss function on quantiles, but can actually be interpreted as a loss function over a cumulative distribution function~$F$ by rewriting it as
\[\ell_{WIS(\tau_{1:M})}(F, y)=\frac{2}{2M+1} \sum_{m=1}^{2M+1} \ell_{\tau_m}(y-F^{-1}(\tau_m)).\]
By increasing the number of quantiles considered, one intuitively expects the WIS loss to be sensitive to any change in the cumulative distribution function $F$.  This intuition can be formalized, as shown in~\cite{Brehmer_2021}, and it can be shown that the WIS loss converges asymptotically to the continuous ranked probability score (CRPS) as $M\to \infty$:
\[\mathrm{CRPS}(F,y) := \int_{-\infty}^{+\infty} \big( F(s) - \mathds 1_{s\geq y}\big)^2 ds.\]
This convergence bridges the gap between quantile-based evaluation and full distributional assessment.

Both the CRPS and WIS are proper scoring rules, i.e. loss functions that are minimized, in expectation, when the forecasted distribution (for CRPS) or the forecasted quantiles (for WIS) match the corresponding true conditional distribution or quantiles  \citep{gneiting2007strictly}. While these proper scoring rules encourages well-calibrated probabilistic forecasts, their adoption in practice remains limited due to several challenges. Firstly, optimizing the risk associated to the CRPS loss can be computationally challenging since explicit solutions are only
available for limited parametric cases \citep{jordan2017evaluating} and gradients of the CRPS are only tractable under some strong assumption on the data distribution \citep{camporeale2021accrue}. Secondly, most existing methods for minimizing the risk associated to the WIS loss require either adding constraints during training or implementing post-processing adjustments to enforce non-crossing quantiles. These additional procedures not only complicate the optimization process but might also negatively impact overall model performance. Thirdly, and perhaps most critically, many of the recent deep learning-based methods for distributional learning, while powerful, suffer from a lack of interpretability. This is a significant drawback in high-stakes applications, such as the medical field, where understanding model decisions is crucial for trust and accountability. There is, therefore, a growing need for interpretable models that can optimize CRPS or WIS losses while maintaining strong predictive performance and computational efficiency.

Among interpretable models, one of the most popular is regression trees. This can be attributed to their intuitive representation of decision-making processes, which appeals to both experts and non-experts. Their versatility in handling diverse data types renders them applicable from finance to healthcare, image recognition, and natural language processing. The construction of decision trees entails an iterative process of selecting features and thresholds to recursively partition the input space, guided by an entropy measure, aka impurity score, quantifying a level of error in each node . At each node, the algorithm assesses potential splitting thresholds for each feature, choosing the one that minimizes the overall impurity. Commonly used impurity measures are in fact directly associated with a convex loss function $\ell$ and  defined as the value of the risk evaluated at its minimizer.  Mathematically, for a loss function $\ell$, the associated impurity measure or generalized entropy $\mathbb H_{\ell}( P)$ is equal to
\begin{equation} \label{eq:entropy_loss}\mathbb H_{\ell}( P) := \inf_{a} \mathbb E_{P}[\ell(a,Y)].\end{equation}
This connection is well-known in the literature \cite[Eq.(2.14)]{degroot62},\citep{grunwald, Aolin}. The generalized entropy associated with a specific loss function is also sometimes referred to as the Bayes risk \citep{Aolin,bao2021fenchel}. In classification, the choice of entropies remains relatively limited. In binary classification, the usual choices for entropy include the Gini entropy, the Shannon entropy, or the 0-1 entropy \citep{generalizedentropy} corresponding respectively to the squared loss, the log loss and the 0-1 loss. For regression, common practice involves opting for a splitting criterion based on the squared loss or the absolute error.

The main contribution of this work lies in the development of an efficient algorithm for learning distributional regression trees that directly optimize the CRPS loss without making any assumptions on the underlying data distribution. To our knowledge, this is the first approach that enables efficient, assumption-free CRPS optimization at scale for trees. In addition, we propose efficient training algorithms for the WIS loss based on an improvement over the approach of \cite{bhat2015}. Our algorithms update multiple conditional quantile estimates concurrently as increasing thresholds on a feature are considered. This makes it easy to define common splits for all quantiles, which produce non-crossing quantile estimates directly. For both CRPS- and pinball-based trees, instead of relying on classical empirical information gain estimates we propose to use better estimates of the population information gain that comes at no extra computational cost, based either on Leave-One-Out (LOO) or, for CRPS-based trees, on a Mallows $C_p$-style correction. This LOO or Mallows criteria penalize complexity and provide a data-driven stopping rule, automatically halting tree growth when splits creating leaf nodes with too few observations do not improve the estimated population risk. Together, these contributions provide practical, scalable, and interpretable methods for calibrated, non-parametric probabilistic forecasting. The proposed methods are particularly relevant for conformal prediction, where a natural choice of prediction sets is based on quantiles of predefined levels \citep{romano2019conformalized} or on estimated conditional distributions \citep{chernozhukov2021distributional}.

\paragraph{Related work.}
To our knowledge, two approaches  to learn a conditional quantile model using regression trees have been proposed in the literature:
\begin{itemize}
\item The first approach is the so-called Quantile Regression Forest (QRF) method \citep{meinshausen2006quantile}.
This algorithm, using the squared loss, benefits from well-established and optimized packages for training regression trees. Additionally, these trees can be used to estimate quantiles of any level with the resulting quantiles guaranteed to be non-crossing (cf. \cite[Section 6]{romano2019conformalized}). However, a key drawback is that the splitting criterion used to construct the tree is well adapted to capture the fluctuations of the conditional mean and therefore not necessarily those of the conditional quantiles of interest, which potentially decreases statistical efficiency.   
\item \cite{bhat2015} consider the problem of learning a single conditional quantile model using a single quantile regression trees trained with the quantile loss function, also known as the pinball loss. 
The entropy associated to the pinball loss is
\begin{align*}\mathbb H_{\ell_{\tau}}( P) &=  \min_{q \in \mathds R} \mathds E_P[ \ell_{\tau} (Y -q)]\\
&= (1-\tau) \mathds E_P[(q-Y)_+] + \tau \mathds E_P[(Y-q)_+],
\end{align*}
where for any real number $x$, $(x)_+=\max(x,0)$.
In the empirical setting where we have observations $\textbf y=\{ y_i \}_{i=1}^n$ sampled from $P$, the empirical entropy is 
\begin{align}\label{eq:empi_pinball} H_{\ell_{\tau}}(\textbf y) :=\min_{q \in \mathds R} \frac{1}{n}\sum_{i=1}^n \ell_{\tau} (y_i -q)=\frac{1}{n}\sum_{i=1}^n \ell_{\tau} (y_i -\hat q_{\tau}),
\end{align}
where $\hat q_{\tau} = y_{(\lceil\tau n\rceil)}$ is the empirical quantile of level $\tau$ of the sequence $\textbf y=(y_i)_{i=1}^n$. The tree is then trained using the standard CART algorithm \citep{breiman2017classification} with the generalized entropy from Eq.\eqref{eq:empi_pinball}: at each node, the data is split by selecting the feature and the threshold that minimize the entropy within the resulting subsets. This process iteratively creates nodes that represent the most homogeneous possible groups by reducing overall entropy.
The procedure to find the data split maximizing the information gain defined with the entropy from Eq.\eqref{eq:empi_pinball} at a given node of the tree with $n$ samples and $d$ features has a time complexity of order $\mathcal O(dn\log n)$ (cf. \cite{bhat2015}). 
\end{itemize}
In this paper, we leverage similar ideas as those of \cite{bhat2015} to propose two efficient algorithms to learn distributional regression trees for the WIS and CRPS losses.

\paragraph{Outline.}

In this work, we first introduce two novel algorithms to learn distributional regression trees (Sec.~\ref{sec:RT}): the first minimizes the empirical entropy associated to the WIS loss, which we call Pinball Multiple Quantiles Regression Trees (PMQRT); the second algorithm learns distributional regression trees (CRPS-RT) for the CRPS loss. To enhance the algorithm, we derive unbiased leave-one-out estimates of the population risk for both PMQRT and CRPS, and further provide a Mallows $C_p$-style correction for CRPS; notably, these corrections incur essentially no additional computational cost (Sec.~\ref{sec:CRPS-LOO}). Detailed algorithmic descriptions, along with illustrative examples, are provided in the supplementary material. In Section~\ref{sec:expes}, we evaluate the empirical performance of random forests of PMQRTs (PMQRF) and CRPS-RTs (CRPS-RF) through a series of experiments. We demonstrate the scalability of the proposed algorithms and that by using a CRPS-RF, we can develop a non-parametric method for learning conditional distribution functions that rivals existing approaches. 

\paragraph{Software Package and Algorithms.}

This paper builds upon several fundamental data structures, namely Weight-Balanced Binary Trees, Fenwick Trees, and Heaps. To assist readers who may be less familiar with these tools, we provide in Section~\ref{apdx:data_structures} concise definitions and concrete examples illustrating how each data structure is used within our algorithms.

All methods presented in this work are available through a publicly released software package called {\it DisTreebution}\footnote{\url{github.com/quentin-duchemin/DisTreebution/}}, which includes complete documentation and usage examples. Section~\ref{sec:package} of the Appendix provides a detailed description of the key features and configurable options of our Python package.

\section{Learning distributional regression trees}
\label{sec:RT}

This section introduces algorithms designed to construct regression trees with common splits for modeling multiple quantiles or the entire conditional distribution. Unlike traditional approaches that treat different quantiles independently, our methods seek to identify shared split points to improve interpretability and computational efficiency. We propose two tree-learning algorithms tailored to distinct loss functions: one for the WIS and another for the CRPS. The following subsections outline these methods and address the challenges inherent in optimizing these distinct objectives.

\subsection{Notations}
To evaluate potential splits based on a single input feature, the vector of output values must first be sorted by increasing order of that input feature. For each possible split index
$s$, we then consider the subset of output values corresponding to the first $s$ elements of the sorted sequence and compute an empirical quantile - that is, an order statistic - of these labels. This motivates the introduction of the following notations: Consider an arbitrary sequence $\textbf a = (a_1,\dots,a_n)$ of real values of size $n$ (here typically label values sorted according to a given input feature). For any $s\in[n]$, we denote by $\textbf a^{(s)}:=(a_1,\dots,a_s)$ the prefix sequence of $\textbf a$. For any $j\in [n]$, $a_{(j)}$ corresponds to the $j$-th largest value of the sequence $\textbf a$, while for any $i \in [s]$, $a^{(s)}_{(i)}$ refers to the $i$-th largest value of the prefix sequence $\textbf a^{(s)}$. 

\subsection{Pinball multi-quantile regression tree}

\label{sec:pmqrts}

\paragraph{Extracting multiple quantiles from a single tree.} 

When estimating multiple quantiles, a natural option is to use QRF, since a single forest can be queried at any level. However, because QRF relies on trees learned with the square loss, it is primarily tailored to conditional mean estimation and may yield suboptimal quantile estimates. Another option is to train a separate tree for each quantile level \(\tau\), following~\cite{bhat2015}, but this does not guarantee non-crossing quantiles and produces a different partition of the feature space for each \(\tau\), which may reduce statistical efficiency and is undesirable for applications such as group conformal prediction (cf. Section~\ref{sec:group_conformel}).



To bypass these limitations, we present an approach that takes the best of the ones above-mentioned. We propose learning a tree that produces all conditional quantiles estimates in each leaf and that most importantly relies on a common split for all quantiles considering the sum of pinball losses $\ell_{\tau_1,\ldots,\tau_m}$ defined for any ${\boldsymbol \tau},\mathbf{q} \in \mathbb{R}^M$ as \[\ell_{\tau_1,\ldots,\tau_m}(\mathbf{q},y):=\sum_{m=1}^M \ell_{\tau_m}(y-q_m).\]
The multi-dimensional optimization problem over $\mathbf{q}\in\mathbb R^M$ that defines the empirical entropy associated with~$\ell_{\tau_1,\ldots,\tau_m}$ is separable, which leads to \begin{equation} \label{eq:multi_entropy}H_{\ell_{\tau_1,\ldots,\tau_m}}(\mathbf{y})=\sum_{m=1}^M H_{\ell_{\tau_m}}(\mathbf{y}).\end{equation}
Relying on the CART algorithm, training a regression tree considering the loss $\ell_{\tau_1,\ldots,\tau_m}$ requires to compute the entropies $H_{\ell_{\tau_1,\ldots,\tau_m}}(\textbf y^{(s)})$ associated with any prefix sequence $\textbf y^{(s)}:= (y_1,\dots, y_s)$ from a given sequence of values $\textbf y=(y_1,\dots,y_n)$. We end up with a problem similar to~\cite{bhat2015} but with multiple quantiles at the same time.

\paragraph{Efficient pinball loss node splitting with max heaps.} 
The algorithm proposed in \cite{bhat2015} to learn a quantile regression tree that minimizes the empirical entropy associated with a single pinball loss $\ell_{\tau}$ uses two heaps (cf. \cite{halim2013competitive}). At a given node in the tree, with data $\{(\textbf{x}_i, y_i)\}_{i=1}^n$, the goal is to find the threshold that provides the best information gain using the entropy from Eq.\eqref{eq:empi_pinball} for each feature $j$. Assuming without loss of generality that $x_{i,j} \leq \dots \leq x_{n,j}$ for some feature $j$, identifying the optimal threshold reduces to computing the entropies for all prefix and suffix sequences of $\textbf{y}$. Since the suffix sequences of $\textbf{y}$ correspond to prefix sequences of $(y_{(n)}, \dots, y_{(1)})$, it is sufficient to explain how \cite{bhat2015} computes efficiently the entropies of the sequences $\textbf{y}^{(s)}$ for all $s \in [n]$. As shown in \cite[Sec.III.B]{bhat2015}, the entropy $H_{\ell_{\tau}}(\textbf{a})$ for any nondecreasing sequence $\textbf{a}$ of length $s$ can be computed in $\mathcal{O}(1)$ time using $a_{\lceil\tau(s-1)\rceil}$, $a_{\lceil\tau s\rceil}$, $\sum_{i=1}^{\lceil\tau(s-1)\rceil} a_{i}$, and $\sum_{i=1}^s a_i$. The authors show that these quantities can be efficiently tracked by initializing two empty heaps - one max-heap $\mathcal H_{low}$ and one min-heap $\mathcal H_{high}$ - and by sequentially adding elements $y_s$ from $s=1$ to $n$ to them. At each step $s$, the update of the two heaps ensures that $\mathcal H_{low}$ contains all elements $(y_{(1)}, \dots, y_{(\lceil\tau(s-1)\rceil)})$ and $\mathcal H_{high}$ contains all elements $(y_{(\lceil\tau s\rceil)}, \dots, y_{(s)})$. Querying the maximum value in $\mathcal H_{low}$ and the minimum value in $\mathcal H_{high}$ can be done in constant time while the heaps update require $\mathcal{O}(\log(s))$ time.

From Eq.~\eqref{eq:multi_entropy}, the empirical entropy of a sum of pinball losses \(\ell_{\tau_1, \dots, \tau_M}\) is the sum of the empirical entropies for each individual pinball loss. As a consequence, a straightforward approach to estimate quantiles at multiple levels using a single tree is to apply the method from \cite{bhat2015} independently for each quantile level. However, this brute-force approach incurs a space complexity of \(\mathcal{O}(Mn)\), making it inefficient as the number of quantile levels \(M\) increases.

To address this limitation, we propose a new algorithm whose space complexity remains of the order of the size of the dataset, regardless of the number of estimated quantiles. 

\paragraph{Efficient pinball multi-quantile loss node splitting with a set of min-max heaps.} 

Using the decomposition Eq.~\eqref{eq:multi_entropy}, one can deduce from~\cite{bhat2015} that the empirical entropy $H_{\ell_{\tau_1,\ldots,\tau_m}}(\textbf y^{(s)})$ of any prefix sequence $\textbf y^{(s)}$ can be obtained by keeping track of $i)$ the empirical quantiles $\widehat q_{\tau_1},\dots,\widehat q_{\tau_M}$ of level $\tau_1,\dots,\tau_M$ of $\textbf y^{(s)}$ and $ii)$ of the cumulative sums $\left( \sum_{i\leq \ceil{\tau_m s}}y^{(s)}_{(i)}\right)_{m=1}^M$ and $\sum_{i\leq s}y^{(s)}_{(i)}$. We propose an efficient algorithm to keep track of these quantities relying on $M+1$ min-max heaps: $\mathcal H_{1},\dots,\mathcal H_{M}, \mathcal H_{M+1}$. Let us recall that a min-max heap is a variant of a binary heap allowing for constant-time access to both the minimum and maximum elements. On even levels, the heap adheres to min-heap properties, while odd levels follow max-heap properties. This ensures that the root always holds the minimum, and its children represent the maximum elements. Insertions and deletions have logarithmic time complexity, providing a balance between efficient extremal element access and effective heap modification.
As for standard regression trees, we increase the index $s$ from 1 to $n$ to cover all the prefix sequences $\textbf y^{(s)}$ of interest to determine the position of the split at the current node. The $M+1$ min-max heaps are updated at each iteration $s$ to ensure that $\mathcal H_{1}$ contains all the elements $\{y^{(s)}_{(1)},\dots y^{(s)}_{(\ceil{\tau_1 s})} \}$, $\mathcal H_{2}$ contains all the elements $\{y^{(s)}_{(\ceil{\tau_1 s}+1)},\dots y^{(s)}_{(\ceil{\tau_2 s})}\}$, $\dots$ , $\mathcal H_{M}$ contains all the elements $\{y^{(s)}_{(\ceil{\tau_{M-1} s}+1)},\dots y^{(s)}_{(\ceil{\tau_M s})}\}$, and $\mathcal H_{M+1}$ contains all the elements $\{y^{(s)}_{(\ceil{\tau_M s}+1)},\dots y^{(s)}_{(s)}\}$. Querying the quantile of level $\tau_m$ simply involves accessing the maximum of the heap $\mathcal H_m$, or the minimum of the heap $\mathcal H_j$ for $j:=\arg \min \{i >m\mid \mathcal H_i\neq \emptyset\}$ in case $\mathcal H_m$ is empty. Our min-max heap structures are also designed to store the total sums of elements in each heap, facilitating access to cumulative sums of the form $\sum_{i\leq \lceil\tau_m s\rceil}y^{(s)}_{(i)}$ in $\mathcal O(m)$ time by querying the total sums of heaps $\mathcal H_1$ to $\mathcal H_m$. Overall, the worst time complexity for a node split with $n$ data points and $d$ features is $\mathcal O(Md n\log n)$ as illustrated in Table~\ref{tab:time_complexity}. The detailed algorithm can be found in Section 2.1 of the supplement.

\subsection{CRPS distributional regression tree}
\label{sec:CRPS}

In the previous section, we presented a computationally efficient method for learning a pinball multi-quantile regression tree. Such a tree has the advantage of leveraging information from different parts of the conditional distribution to build the partition, ensuring non-crossing quantile estimates, and offering a flexible tool for modern uncertainty quantification, ideal for applications such as group conformal methods, see Section~\ref{sec:group_conformel}. However, a Pinball multi-quantile regression tree focuses only on specific quantiles, limiting its ability to represent the full conditional distribution. As a result, it may overlook key aspects of the data distribution, affecting predictive accuracy. Furthermore, certain methods, such as the distributional conformal prediction method proposed in \cite{chernozhukov2021distributional}, require querying quantiles at arbitrary levels, and constraining the model to a fixed set of predefined quantiles may hinder its effectiveness. 

To bypass these limitations, we propose to learn trees that minimize the empirical entropy associated with the Continuous Ranked Probability Score (CRPS), namely \begin{equation}  H_{\mathrm{CRPS}}(\textbf y):=\min_F \frac1n \sum_{i=1}^n \mathrm{CRPS}(F,y_i) = \frac1n \sum_{i=1}^n \mathrm{CRPS}(\widehat F_n,y_i) , \label{eq:CRPS_empirical}\end{equation}
where $\widehat F_n$ is the empirical CDF of the observed data $\textbf y$. The CRPS is a strictly proper scoring rule, as detailed in \cite{gneiting2007strictly}, incentivizing forecasters to provide accurate and well-calibrated uncertainty estimates, making it a valuable tool in probabilistic forecasting evaluations.

\paragraph{CRPS-Based tree training: Theoretical foundations and design.}

To construct a distributional regression tree based on the generalized entropy induced by the CRPS loss, one must identify the split at a given leaf that results in the largest impurity gain. 
Given a leaf with data $\left\{(\textbf{x}_i, y_i)\right\}_{i=1}^n$, this process involves computing the entropies for every feature $j$ using every prefix sequence $\textbf{y}^{(s)} = \left\{y_1, \dots, y_s\right\}$ and every suffix sequence $(y_{s+1}, \dots, y_n)$ of the data sequence $\textbf{y}$, assuming for simplicity that $x_{1,j} \leq \dots \leq x_{n,j}$. Relying on the entropy expression from Eq.~\eqref{eq:CRPS_empirical}, this requires constructing the empirical CDFs for all prefix and suffix sequences of $\textbf{y}$ for each feature $j$. This process is computationally intensive: specifically, the construction of empirical CDFs for the prefix sequences of a feature $j$ requires sequentially inserting a new element $y_{s+1}$ into the current sorted version of the list $\textbf{y}^{(s)}$, with a time complexity of $\mathcal{O}(s)$. Consequently, the basic construction of a distributional regression tree based on the CRPS loss results in a node split with a time complexity of $\mathcal{O}(d n \log(n) \sum_{i=1}^n s) = \mathcal{O}(d n^3 \log n)$, which is not feasible for practical use.

To efficiently train a tree using the CRPS loss, reducing computations is key. Due to the sequential structure of the computation of entropies associated with prefix (resp. suffix) sequences to perform a node split, one can investigate the possibility to rely on iterative schemes to reduce the computational burden of the algorithm. Theorem~\ref{thm:indu} presents an iterative formula allowing us to compute the entropies associated to all prefix sequences for a given feature $j$. Note that a similar formula is directly obtained from Theorem~\ref{thm:indu} for suffix sequences by flipping the sign of each entry of the feature vectors $\left\{\textbf x_i\right\}_{i=1}^n$. In what follows, we denote by $h^{(s)}_{\text{mirror}}$ the counterpart of the entropies $h^{(s)}$ from Theorem~\ref{thm:indu} computed under this sign-flipped transformation of the input features.

\begin{theorem}
 Consider a leaf with data $\left\{(\textbf{x}_i, y_i)\right\}_{i=1}^n$ and a given feature $j$. Assume without loss of generality that $x_{1,j} \leq \dots \leq x_{n,j}$ and let us denote by $\textbf y^{(s)}$ the prefix sequence $(y_1,\dots,y_s)$ of size $s$ and $\widehat F_s$ the empirical CDF of $\textbf y ^{(s)}$. Then, the empirical entropies 
\[ H_{\mathrm{CRPS}}(\textbf y^{(s)}):= \frac{1}{s} \sum_{i=1}^s \mathrm{CRPS}(\widehat F_s, y_i),\]
for all $s\in[n]$ can be computed using an iterative scheme. Indeed, we have
\[ H_{\mathrm{CRPS}}(\textbf y^{(s+1)})=\frac{1}{(s+1)^3}\left(s^3 H_{\mathrm{CRPS}}(\textbf y^{(s)}) + h^{(s+1)}\right),\]
with $h^{(0)}=0$ and for any $s\geq 1$,
\begin{align*}
h^{(s)}
&=h^{(s-1)}+(s-1)\big[2r_{s} -s\big] y^{(s)}_{(r_{s})} + 2 W^{(s)}-2S^{(s)}-2(s-1)S^{(s)}_{r_{s}},
\end{align*}
and where for any $r\in[s]$, 
\[S^{(s)}_r:=\sum_{i=1}^{r}y^{(s)}_{(i)}, \quad W^{(s)}:=\sum_{i=1}^s iy^{(s)}_{(i)} \quad \text{and}\quad S^{(s)}:=S^{(s)}_s,\]
and, for any $s\in[n-1]$, $r_{s+1}$ is the insertion rank of $y_{s+1}$ within the non-decreasing sorted version of the list $\textbf y^{(s)}$.   
\label{thm:indu}
\end{theorem}

Theorem~\ref{thm:indu} shows that training a distributional regression tree using the CRPS loss is achievable if the insertion ranks \(\big(r_s\big)_{s \in [n]}\) and cumulative sums \(S^{(s)}_{r_s}\), \(S^{(s)}\), and \(W^{(s)}\) can be computed efficiently. To this end, we introduce the \textit{CRPS-Optimal Split} algorithm, which leverages tailored data structures to compute the entropy of all candidate splits for any feature $j$ in $\mathcal O(n \log n)$ time. The procedure is presented in Algorithm~\ref{alg:crps_split} and described in detail in the remainder of this section. 
To efficiently compute the ranks \( r_s \), the algorithm uses a Weight-balanced tree. This is a type of self-balancing binary search tree that keeps its shape roughly balanced as elements are inserted or removed. Because of this balance, operations like insertion, deletion, and searching can all be done in \( \mathcal{O}(\log n) \) time.

In our case, we use the Weight-balanced tree to determine the rank of each element as it is inserted — that is, how many elements in the current set are smaller than the new one. This can be viewed as inserting each element into a dynamically maintained sorted list while simultaneously determining its sorted position (rank), without requiring a full sort of the list. This rank computation is also performed in \( \mathcal{O}(\log n) \) time thanks to the structure of the tree.

After computing the ranks $(r_s)_s$, we use a Fenwick tree (also known as a binary indexed tree) to compute the cumulative sums \( S^{(s)}_{r_s} \). A Fenwick tree is a clever data structure that allows us to efficiently update values and calculate prefix sums — that is, the sum of values up to a certain index. Unlike a regular array, which would need up to \( n \) steps to compute a prefix sum, a Fenwick tree can compute it in just \( \mathcal{O}(\log n) \) time. It achieves this by storing partial sums in a compact binary representation, enabling efficient updates and queries. An illustration of how the Fenwick Tree works is shown in Figure~\ref{fig:fenwick_tree}.

\begin{figure}[!ht]
\centering
\begin{tikzpicture}[level distance=2.5cm,
level 1/.style={sibling distance=3.5cm},
level 2/.style={sibling distance=2.3cm}]
\tikzstyle{every node}=[rectangle,draw]

\node[text width=1.8cm,align=center] (Root) [red] {0 : 0b000\\ $\sum_{i=0}^0 a_i$}
    child {
    node[text width=1.8cm,align=center] {1 : 0b001 \\ $\displaystyle \sum_{0<i\leq 1} a_i$}
    }
    child {
    node[text width=1.8cm,align=center] {2 : 0b010 \\ $\displaystyle \sum_{0<i\leq 2}  a_i$}
    child { node[text width=1.8cm,align=center] {3 : 0b011 \\ $\displaystyle \sum_{2<i\leq 3}  a_i$} }
    }
    child {
    node[text width=1.8cm,align=center] {4 : 0b100 \\ $\displaystyle \sum_{0<i\leq 4}  a_i$}
    child {  node[text width=1.8cm,align=center] {5 : 0b101 \\ $\displaystyle \sum_{4<i\leq 5}  a_i$} }
    child { node[text width=1.8cm,align=center] {6 : 0b110 \\ $\displaystyle \sum_{4<i\leq 6} a_i$}  
        child {   node[text width=1.8cm,align=center] {7 : 0b111 \\ $\displaystyle \sum_{6<i\leq 7}  a_i$} }
        }
    };
\end{tikzpicture}
 \caption{Fenwick tree for a list $(a_i)_{i\in\{1,\dots,7\}}$ of length $7$. By convention, $a_0=0$. The nodes of the tree are indexed by the natural integers. $0$ is at the root and the children of the root correspond to all integers of the form $2^p$ with $p\in \mathbb{N}.$ More generally, nodes at depth $k$ are the sum of $k$ integers of the form $2^p,$ i.e., they have $k$ non-zero bits in their binary expansion. The parent of node $i$ has the same binary expansion except that the lowest non-zero bit of $i$ is set to $0$ in its parent. If $A_j=\sum_{i=0}^j a_i,$ then the partial sum stored at node $i$ is $A_i-A_{\texttt{pa}(i)},$ where $\texttt{pa}(i)$ is the parent of $i$. }
\label{fig:fenwick_tree}
\end{figure}

\begin{algorithm}
\caption{CRPS-Optimal Split: Optimal CRPS Node Split for feature $j$}
\begin{algorithmic}[1]
\State \textbf{Input:} Feature values $\mathbf{x}_{:,j}$ and target vector $\mathbf{y} = (y_1, \dots, y_n)$, sorted such that $x_{1,j} \leq \dots \leq x_{n,j}$
\State Let $\sigma$ be the permutation sorting $\mathbf{y}$ in non-decreasing order: $y_{\sigma(1)} \leq \dots \leq y_{\sigma(n)}$
\State Initialize: $W^{(0)} \gets 0$, $h^{(0)} \gets 0$
\State \textbf{For} $s = 1$ to $n$ \textbf{do}:
\vspace{0.3em}
\begin{center}
\renewcommand{\arraystretch}{1.4}
\begin{tabular}{|>{\centering}p{4.9cm}|>{\centering}p{5.3cm}|>{\centering\arraybackslash}p{2.5cm}|}
\hline
\textbf{Computation} & \textbf{Operation \& Data Structure} & \textbf{Time Complexity} \\
\hline
Compute rank $r_s$ & Query rank via inserting $y_s$ in a Weight-balanced tree & $\mathcal{O}(\log n)$ \\
\hline
Compute partial sum $S_{r_s}^{(s)} = \sum_{i=1}^{r_s} y_{(i)}^{(s)}$ & Insert the value $y_s$ and query the prefix sum from 1 to $\sigma^{-1}(s)$ in the Fenwick tree & $\mathcal{O}(\log n)$ \\
\hline
Update weighted sum $W^{(s)}$ & Incremental update from $W^{(s-1)}$ and $S_{r_s}^{(s)}$ & $\mathcal{O}(1)$ \\
\hline
Update entropy $h^{(s)}$ & Closed-form update using $W^{(s)}$, $h^{(s-1)}$, and $S_{r_s}^{(s)}$ & $\mathcal{O}(1)$ \\
\hline
\end{tabular}
\end{center}
\vspace{0.3em}
\State Repeat the above loop with $\mathbf{y}$ in reverse order to obtain $h^{(s)}_{\text{mirror}}$
\State \textbf{Output:} Optimal split index $s^*(j)$ and the corresponding entropy
\[
s^*(j) = \arg\min_s \left( s \cdot h^{(s)} + (n - s) \cdot h^{(s)}_{\text{mirror}} \right)
\]
\end{algorithmic}
\label{alg:crps_split}
\end{algorithm}

\paragraph{CRPS-Based tree training: Detailed procedure.}
For a given feature $j$, the process begins by sorting the list of feature entries $\left\{x_{i,j}\right\}_{i=1}^n$, which requires $\mathcal{O}(n \log n)$ time. Once sorted, the insertion ranks $\big(r_s\big)_{s \in [n]}$ can be pre-computed using a weight-balanced tree \citep{Nievergelt73}. Weight-balanced trees offer logarithmic performance similar to that of red-black and AVL trees but also provide the advantage of indexing elements in logarithmic time. Lemma~\ref{lemma:WB_tree} shows that these ranks can be pre-computed in $\mathcal{O}(n \log n)$ time.

\begin{lemma} \cite[Chapter 10]{mehta2004handbook}
A weight-balanced tree with data $\textbf{y}^{(s)}$ allows inserting a new value $y_{s+1}$ and determining its position in the sorted version of the list $\textbf{y}^{(s)}$ such that the resulting list remains in non-decreasing order, all in $\mathcal O(\log s)$ time.
\label{lemma:WB_tree}
\end{lemma}

Once the ranks $\big(r_s\big)_{s \in [n]}$ are computed, we need to obtain the entropies $ H_{\mathrm{CRPS}}(\textbf y^{(s)})$ for any prefix sequence $\textbf y^{(s)}$ using the iterative scheme presented in Theorem~\ref{thm:indu}. For this, it is sufficient to query efficiently the sums $S^{(s)}_{r_s}$, $S^{(s)}$ and $W^{(s)}$. 
Tracking \( S^{(s)} = \sum_{i=1}^s y_i \) across iterations is straightforward, as it simply involves adding the new element \( y_s \) to \( S^{(s-1)} \) at each step \( s \). Moreover, we observe that  
\begin{equation*}\label{eq:key_update_CRPS}W^{(s+1)} = W^{(s)} +r_{s+1}y_{s+1} +\sum_{i=r_{s+1}}^s y^{(s)}_{(i)}=W^{(s)}  +(1+r_{s+1})y_{s+1} +S^{(s)}-S^{(s+1)}_{r_{s+1}},\end{equation*}
where we used that $\sum_{i=r_{s+1}}^s y^{(s)}_{(i)}=\sum_{i=r_{s+1}}^{s+1} y^{(s+1)}_{(i)}-y_{r_{s+1}}^{(s+1)}=S^{(s+1)}_{r_{s+1}}-y_{s+1}$.
Thus, maintaining \( W^{(s)} \) comes at no additional cost if partial sums of the form \( S^{(s)}_{r_s} \) can be queried efficiently. This highlights that efficiently tracking these sums is the key to designing an algorithm that constructs regression trees for the CRPS loss. 
Denoting by $\sigma$ a permutation sorting the list $\textbf y$ in non decreasing order and having in mind that $y_s=y^{(s)}_{(r_s)}$, the crucial observation is:
\begin{equation}\label{eq:motiv_fenwick}S^{(s)}_{r_s}=\underbrace{\sum_{i=1}^{r_s} y^{(s)}_{(i)}}_{\text{$r_s$ smallest elements of $\mathbf y^{(s)}$}} = \underbrace{\sum_{i=1}^s \mathrm 1_{\sigma^{-1}(i)\leq \sigma^{-1}(s)} y_i}_{\text{Elements of $\mathbf y^{(s)}$ smaller than $y_s$}}=\underbrace{\sum_{j=1}^{\sigma^{-1}(s)} \mathrm 1_{\sigma(j)\leq s} y_{\sigma(j)}}_{\substack{\text{Elements of $\textbf y$ smaller}\\\text{than $y_s$ with index$ \leq s$}}}.\end{equation}
Building on Eq.\eqref{eq:motiv_fenwick}, we rely on a customized Fenwick tree to efficiently track the sums $S^{(s)}_{r_s}$. A Fenwick tree, as shown in Figure~\ref{fig:fenwick_tree}, is a binary-indexed data structure that efficiently supports prefix sum and insertion in logarithmic time. In Definition~\ref{def:fenwick}, we introduce the tailored Fenwick tree $\mathcal{F}$, which serves as the cornerstone of our algorithm, enabling efficient node splits.

\begin{definition}
Let us denote by $\sigma$ a permutation sorting the list $\textbf y$ in non decreasing order (i.e. $y_{\sigma(1)}\leq \dots \leq y_{\sigma(n)}$). We define by $\mathcal F$ a Fenwick tree of size $n$ initialized with $n$ zero values such that at any iteration $s\geq 1$, we add the element $y_s$ at the position $\sigma^{-1}(s)$ in the list represented by the Fenwick tree $\mathcal F$. 
\label{def:fenwick}
\end{definition}
 By construction at any iteration $s$, $\mathcal F$ is a Fenwick tree for the list $(\mathrm 1_{\sigma(i)\leq s} y_{\sigma(i)} )_{i\in [n]}$.
In Lemma~\ref{lemma:fenwick_tree}, we demonstrate that $\mathcal F$ enables tracking the cumulative sums of the form $S^{(s)}_{r_s}$ and $S^{(s)}$ which, as a by product, allows 
 for monitoring $W^{(s)}$ throughout the iterative process. The proof of Lemma~\ref{lemma:fenwick_tree} is provided in Section 3.2 of the supplement. 

\begin{lemma}
At iteration $s$, the structure of the Fenwick tree $\mathcal F$ makes it possible to:
\begin{itemize}
    \item calculate $S^{(s)}_{r_s}$ and $S^{(s)}$ in $\mathcal{O}(\log n)$ time,
    \item insert a new value in $\mathcal{O}(\log n)$ time,
    \item compute $W^{(s)}$ from $W^{(s-1)}$ in $\mathcal{O}(\log n)$ time.
\end{itemize}
\label{lemma:fenwick_tree}
\end{lemma}

In Section 2.2 of the supplement, we present in details our algorithm to perform a node split to grow a CRPS-RT. Overall, the computational time required to perform a node split scales as $\mathcal O(dn\log n) $ where $n$ is the number of data points in the node and $d$ is the number of features. The time complexity of the two algorithms presented in this paper are given in Table~\ref{tab:time_complexity}.

\begin{table}[!ht]
\centering
\begin{tabular}{c|C{3cm}||C{3cm}}
\textbf{Algorithm} & Multiple Quantiles Regression Tree & CRPS Regression Tree\\\hline
\textbf{Time complexity} & $\mathcal O(Mdn\log n )$ & $\mathcal O(dn\log n ) $
\end{tabular}
\caption{Worst case time complexity for a specific node split for the different regression trees.}
\label{tab:time_complexity}
\end{table}

\subsection{Estimating the population entropies }
\label{sec:CRPS-LOO}

To train PQRTs, PMQRTs or CRPS-RTs, we proposed in the previous sections to rely on the standard CART algorithm \citep{breiman1984classification} which is based on empirical entropies. Unfortunately, these are biased estimates of the population entropies (i.e., associated with the population risk), because the same data is used both to estimate the empirical quantiles or empirical cumulative distribution functions and to evaluate the corresponding in-sample empirical risk. As a result, the empirical criterion computed at each step of the tree-growing procedure is typically an optimistically biased estimate of the target prediction risk, and therefore induces a biased estimate of the associated information gain. A naive way to reduce this bias would be to split the sample into two disjoint parts, one used for estimation and the other for evaluation. Such a strategy is, however, statistically inefficient, since only a fraction of the available observations is used for each task at every candidate split.

To address this issue, we consider two complementary bias-correction strategies: leave-one-out (LOO) and a Mallows $C_p$-style optimism correction. Using the LOO or Mallows estimate of the information gain provides a data-driven stopping criterion: if all corrected information gains are non-positive, the current node is not split. In contrast, this situation can only arise in trivial settings when using the standard information gain estimate, because its optimistic bias ensures that the gain values are always non-negative.

Despite their conceptual appeal, the use of LOO and Mallows corrections in regression trees could, in principle, introduce a computational overhead. Indeed, applying these corrections at each candidate split could be costly, since the information gain would need to be evaluated for all possible splits when determining the best node split. In this section, we show that a LOO (resp. Mallows) estimate of the information gains can be obtained at no additional computational cost for the WIS and CRPS losses (resp. CRPS loss). Before presenting the formula for computing these estimates, we make a brief and subtle clarification regarding the distinction between the LOO and Mallows targets.

\paragraph{LOO vs Mallows $C_p$-style estimate of the information gains.}

The targets of the LOO and Mallows estimates differ slightly. Consider a node containing $n$ observations $(y_1,\dots,y_n)$ and let $\hat f_{k:j}^{\mathrm{ERM}}$ denote the empirical risk minimizer for the loss $\ell$ fitted on the subsample $(y_k,\dots,y_j)$. In the PMQRT setting, $\hat f_{k:j}^{\mathrm{ERM}}$ is the vector of empirical quantiles associated with the prescribed quantile levels, whereas in the CRPS-RT setting it is the empirical cumulative distribution function of the subsample. We write
\[
\widehat R_{k:j}(\hat f_{k:j}^{\mathrm{ERM}}):=H_\ell(\mathbf y_{k:j})
\]
for the corresponding empirical risk (or impurity), and $R(\hat f_m^{\mathrm{ERM}})$ for the population risk of the same learning procedure trained on $m$ observations. For a candidate split producing children of sizes $k$ and $n-k$, a Mallows $C_p$-style correction directly targets the expected information gain associated with the full-sample models, namely
\[
\mathbb E\!\left[R(\hat f_n^{\mathrm{ERM}})-\bigl(R(\hat f_k^{\mathrm{ERM}})+R(\hat f_{n-k}^{\mathrm{ERM}})\bigr)\right],
\]
by correcting the empirical information gain with the appropriate optimism terms. By contrast, LOO evaluates each node with one observation removed and therefore targets the analogous quantity for models trained on $n-1$, $k-1$, and $n-k-1$ observations. Hence, LOO is not exactly unbiased for the target full-sample information gain, although the discrepancy is typically small when node sizes are large. A more detailed discussion is provided in the supplementary material. 
Our numerical experiments described in Section~\ref{sec:LOO_Mallows_expes} have shown that building trees using either LOO estimates of the information gains or a Mallows correction leads to very consistent results.


\paragraph{LOO estimate of the information gains.}
For both the CRPS loss and the sum of pinball losses, we derive a LOO correction that yields an unbiased estimate of the corresponding empirical entropy while remaining computationally cost-free. In particular, the naive implementation of LOO would appear prohibitively expensive, since it requires recomputing the impurity criterion after removing each observation in turn. Nevertheless, Propositions~\ref{prop:LOO_CRPSRT} and~\ref{prop:LOO_PMQRT} show that, in both cases, the resulting LOO criterion admits a simple closed-form expression and can be evaluated with no additional computational overhead.

\begin{proposition}
\label{prop:LOO_CRPSRT}
Let $H_{\mathrm{CRPS}}$ denote the entropy of $\textbf{y}$ for the CRPS loss and $H_{\mathrm{CRPS}}^{LOO}$ its LOO version, defined as:
\[
H_{\mathrm{CRPS}}(\textbf y)= \frac{1}{n} \sum_{i=1}^{n} \mathrm{CRPS}(\widehat F_n,y_i), \quad H_{\mathrm{CRPS}}^{LOO}(\textbf y) = \frac{1}{n} \sum_{i=1}^{n} \mathrm{CRPS}(\widehat F_n^{-i}, y_i),
\]
where $\widehat F_n$ and $\widehat F_n^{-i}$ represent the empirical CDF of $\textbf{y}$ and~$\textbf{y}^{-i} = (y_1, \dots, y_{i-1}, y_{i+1}, \dots, y_n)$, respectively. We have
    \[
    H^{LOO}_{\mathrm{CRPS}}(\textbf y)=\frac{n^2}{(n-1)^2}  
        H_{\mathrm{CRPS}}(\textbf y).
    \]
\end{proposition}

Proposition~\ref{prop:LOO_PMQRT} shows that the LOO estimate of empirical entropy for the pinball loss, \( H_{\ell_{\tau}}^{\mathrm{LOO}}(\textbf{y}) \), can be derived from \( H_{\ell_{\tau}}(\textbf{y}) \) by adding a debiasing term that depends on \( y_{(r^*-1)} \), \( y_{(r^*)} \), and \( y_{(r^*+1)} \), where \( r^* := \lceil n\tau \rceil \). Since these values can be retrieved in constant time using the min-max heap structures described in Section~\ref{sec:pmqrts},  a node split, whether using LOO or not, is still performed in 
 $\mathcal O(dMn \log n)$ time when considering a sum of $M$ pinball losses. The proofs of Propositions~\ref{prop:LOO_CRPSRT} and~\ref{prop:LOO_PMQRT} are provided in the supplement.

\begin{proposition}
\label{prop:LOO_PMQRT}
Let $H_{\ell_{\mathbf \tau}}$ denote the entropy of $\textbf{y}$ for the pinball loss $\ell_{\tau}$ and $H_{\ell_{\mathbf \tau}}^{LOO}$ its LOO version, defined as:
\[
H_{\ell_{\mathbf \tau}}(\textbf y)= \frac{1}{n} \sum_{i=1}^{n} \ell_{\tau}(y_i, \hat{q}_{\tau}), \quad H_{\ell_{\mathbf \tau}}^{LOO}(\textbf y) = \frac{1}{n} \sum_{i=1}^{n} \ell_{\tau}(y_i, \hat{q}_{\tau}^{-i}),
\]
where $\hat{q}_{\tau}$ and $\hat{q}_{\tau}^{-i}$ represent the empirical quantiles of order $\tau$ computed over $\textbf{y}$ and~$\textbf{y}^{-i} = (y_1, \dots, y_{i-1}, y_{i+1}, \dots, y_n)$, respectively. Setting $r^- := \lceil \tau (n-1) \rceil$ and $r^* := \lceil n\tau \rceil$, we have
    \[
    H^{LOO}_{\ell_{\mathbf \tau}}(\textbf y)= \left\{
    \begin{array}{ll}
        H_{\ell_{\mathbf \tau}}(\textbf y) + \frac{(1-\tau) r^*}{n} (y_{(r^*+1)} - y_{(r^*)})  & \mbox{if } r^*=r^- \\
        H_{\ell_{\mathbf \tau}}(\textbf y) +\frac{\tau (n - r^* + 1)}{n} (y_{(r^*)} - y_{(r^*-1)}) & \mbox{if }r^*=r^- +1.
    \end{array}
\right.
    \]
\end{proposition}

\paragraph{Mallows-style correction for the information gains.}
In the specific case of the CRPS loss, we additionally derive a Mallows $C_p$-style optimism correction. As presented in Proposition~\ref{prop:Mallows_CRPSRT} (see the Supplement for the proof), this correction can be computed at essentially no additional computational cost and provides an unbiased estimate of the CRPS information gain at each candidate split. 
By contrast, for the sum of pinball losses, no computationally cheap Mallows-type correction is currently available to the best of our knowledge. While a Mallows-style correction for the pinball loss might exist in principle, deriving one without introducing substantial additional computational overhead appears substantially more challenging, and we therefore leave this question for future work.

\begin{proposition}
\label{prop:Mallows_CRPSRT}
 The Mallows $C_p$-type corrected version of the empirical entropy $H_{\mathrm{CRPS}}(\textbf y)$ is given by
\[
H_{\mathrm{CRPS}}(\textbf y)+\frac{2}{n-1}H_{\mathrm{CRPS}}(\textbf y)
=
\frac{n+1}{n-1}H_{\mathrm{CRPS}}(\textbf y),
\]
and provides an unbiased estimator of the population CRPS risk $\mathbb E[R(\widehat F_n)]$.
\end{proposition}



\section{Experiments}
\label{sec:expes}
In this section, we investigate the empirical performance of the different algorithms introduced in this paper when they are used within random forests based on the principle of bagging. In a first experiment, we assess the usefulness of the LOO and Mallows estimates of the population entropies. In a second part, we study the empirical computational efficiency of Algorithm~\ref{alg:crps_split}. In a third part, we compare our methods with several alternative approaches, in particular QRF \citep{meinshausen2006quantile}. Finally, we present an application of the distributional regression trees introduced in this work to group conformal prediction.

When using trees, there are two natural aggregation schemes, depending on whether the primary focus is on conditional quantiles or on conditional CDFs. Given a new input $x$, a first approach follows the aggregation strategy of QRF, where a conditional CDF $\hat F(\cdot \mid X=x)$ is obtained by averaging the empirical CDFs $ \hat{F}_k(\cdot \mid X = x)$ produced by each tree $k$ in the forest. More precisely, for each tree $k$, one considers the empirical CDF of the observations contained in the leaf reached by dropping $x$ from the root of the tree, and defines
\[
\hat{F}(y \mid X = x) = \frac{1}{K} \sum_{k=1}^K \hat{F}_k(y \mid X = x).
\]
We refer to this first approach as \emph{Distributional Bagging} (DB).

A second approach consists in averaging the conditional quantiles produced by each tree. Specifically, for each tree $k$, let $\hat{q}_{k,\beta}(Y \mid x)$ denote the empirical quantile of level $\beta$ computed from the observations in the leaf reached by dropping $x$ from the root of tree $k$. One then defines
\[
\hat{q}_{\beta}(Y \mid x) = \frac{1}{K} \sum_{k=1}^K \hat{q}_{k,\beta}(Y \mid x).
\]
We refer to this second approach as \emph{Quantile Bagging} (QB).

Although DB is naturally formulated as an aggregation scheme for conditional CDFs and QB as an aggregation scheme for conditional quantiles, both approaches can in fact be used to recover either object. In particular, DB directly provides an estimate of the conditional CDF $\hat F(\cdot \mid X=x)$, from which conditional quantiles can be obtained by querying $\hat F$. Conversely, QB directly provides conditional quantile estimates as a function of the level $\beta$, and a conditional CDF can be recovered by considering the corresponding stepwise inverse representation induced by the collection of conditional quantiles.

Both aggregation schemes were implemented and compared in our experiments. In all experiments involving PMQRF or CRPS-RF, we used the QB aggregation scheme. This is motivated by preliminary experiments that show that QB performs comparably or slightly better (cf. Section~\ref{apdx:study_agg}).

\subsection{CDF estimation improvement with an unbiased CRPS risk estimate}
\label{sec:LOO_Mallows_expes}

\begin{figure*}[!ht]
    \centering
    \begin{subfigure}[t]{0.5\textwidth}
        \centering
        \includegraphics[height=2.2in]{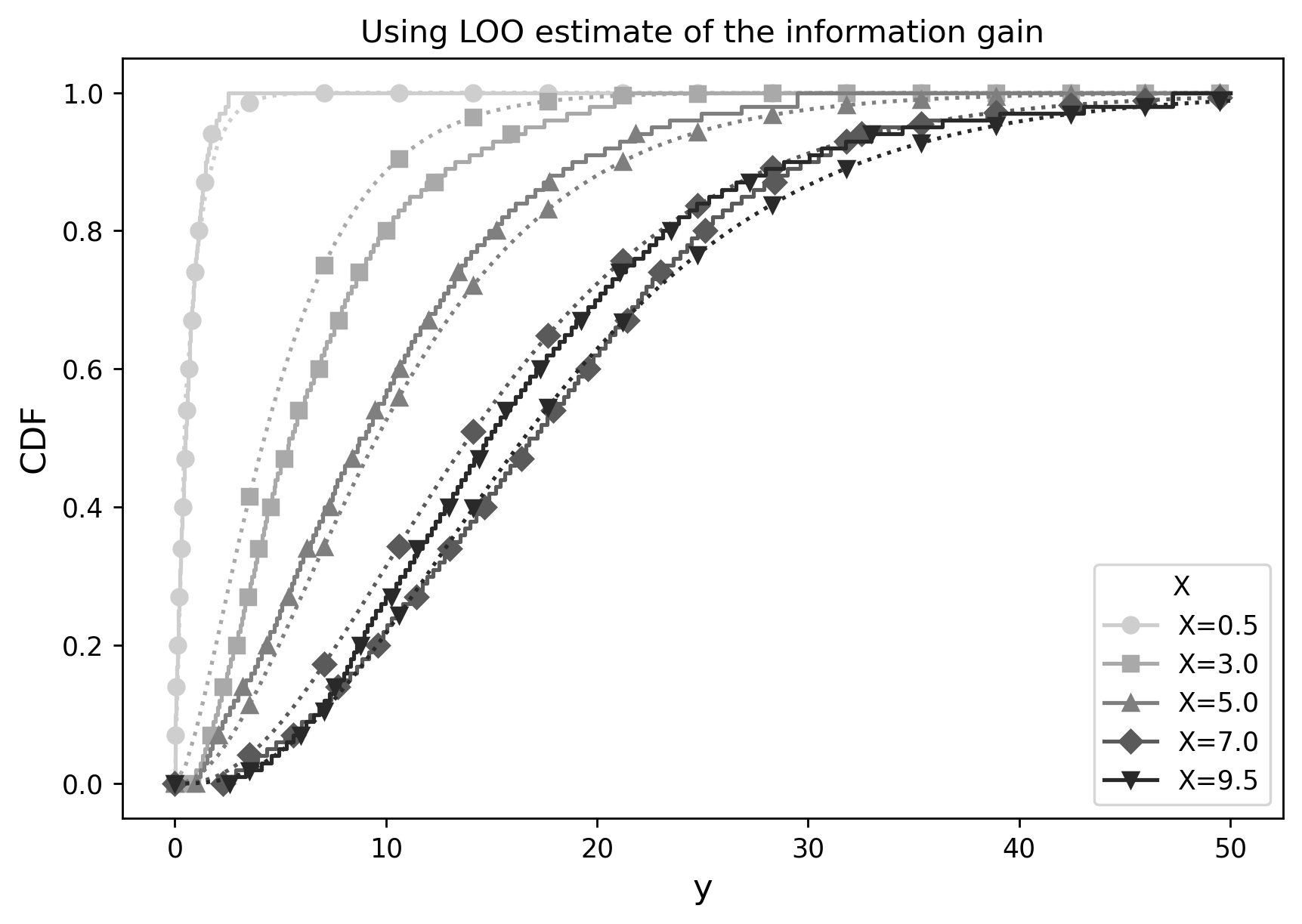}
        \caption{With LOO.}
    \end{subfigure}%
    ~ 
    \begin{subfigure}[t]{0.5\textwidth}
        \centering
        \includegraphics[height=2.2in]{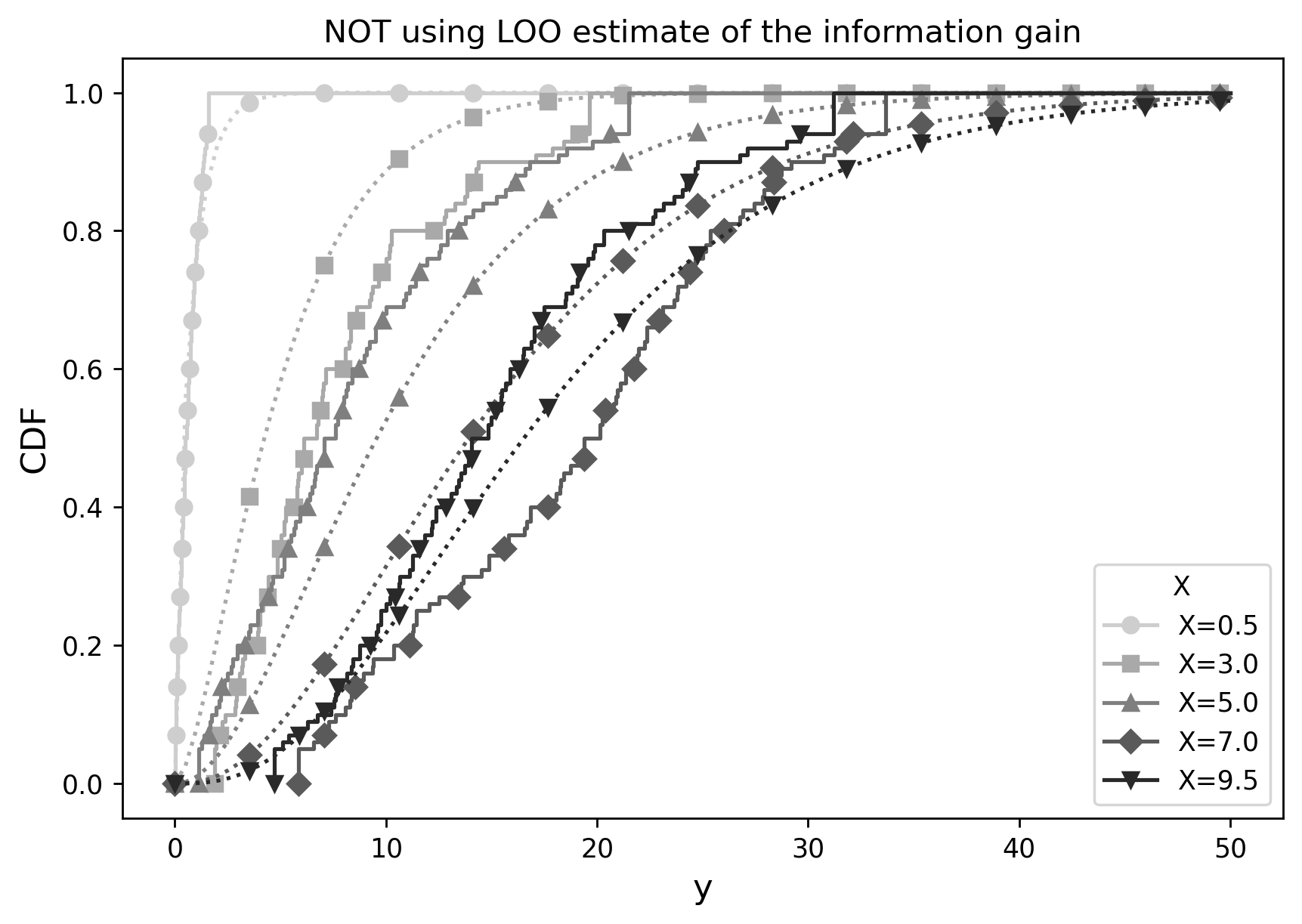}
        \caption{Without LOO.}
    \end{subfigure}
    \caption{Estimation of the conditional CDF of y given x with CRPS-RF on a sample of size $n=600$ simulated from the distribution in Eq.\eqref{eq:simu_gamma}. The true conditional CDFs (smooth dashed lines) are compared to the estimates obtained from CRPS-RF (solid line step functions).}
    \label{fig:simu_LOO}
\end{figure*}

To demonstrate the importance of using LOO for calculating information gain, we consider a simple example where the covariate $X$ is uniformly sampled from the interval $(0, 10)$, and \begin{align} \label{eq:simu_gamma} Y \mid X &\sim \text{Gamma}(\text{shape} = \sqrt{X}, \text{scale} = \min\{\max\{X, 1\}, 6\})
\end{align}
Figure~\ref{fig:simu_LOO} shows the estimated conditional CDFs using a forest of 100 CRPS-RTs, comparing the results with and without LOO used for entropy calculation. On this example, applying LOO produces a significantly better estimate of the conditional CDFs. We perform a similar experiment for PMQRF, demonstrating a significant improvement in quantile estimates achieved through the use of LOO. The results of these simulations are presented in Section 4 of the supplement.

We extend the analysis by quantifying the statistical implications of using LOO and Mallows-style corrections to compute the information gain. We consider CRPS-RF models trained with 100 trees on $600$ data points sampled from Eq.~\eqref{eq:simu_gamma}, exploring various values of the stopping-criterion hyperparameters $max\_depth$ (maximum depth of trees) and $min\_samples\_split$ (the minimum number of samples required to split an internal node).  For each configuration, we train three versions of CRPS-RF: one using LOO, one using the Mallows $C_p$-style correction, and one without any correction.

Once trained, we query the estimated quantiles $\hat{q}_{0.05}$ and $\hat{q}_{0.95}$ (corresponding to the 5\% and 95\% quantiles, respectively) on a test set of 1,000 data points, and compute the marginal coverage induced by the predictive intervals $[\hat{q}_{0.05}, \hat{q}_{0.95}]$. The results are reported in Table~\ref{tab:LOO_coverage}.

Empirically, when the stopping rules are sufficiently restrictive (i.e., small $max\_depth$ and/or large $min\_samples\_split$), the marginal coverages of CRPS-RF models trained with LOO, Mallows, or without correction are comparable. Conversely, when the stopping rules are less restrictive (i.e., large $max\_depth$ and small $min\_samples\_split$), models trained without correction tend to exhibit marginal coverages that deviate more strongly from the target level of 90\%, whereas models trained with either LOO or Mallows maintain much more stable coverage.

In the absence of correction, tree growth generally continues until a predefined stopping criterion is satisfied, since each node split typically produces a positive information gain, except in trivial cases. In contrast, both LOO and Mallows corrections yield improved estimates of the population entropies. Once the corrected information gains become non-positive for all possible splits of a node, the splitting process naturally terminates, even if the explicit stopping criteria have not been reached. Consequently, incorporating either LOO or Mallows-type corrections effectively implements a data-driven, automatic stopping rule.

\begin{table}[h!]
\centering
\caption{Comparison of mean coverage between no correction, LOO, and Mallows across $max\_depth$ and $min\_samples\_split$.}
\label{tab:LOO_coverage}
\scriptsize
\setlength{\tabcolsep}{3pt}
\renewcommand{\arraystretch}{1.1}
\begin{tabular}{|c|c|c|c|c|c|c|c|c|c|c|c|c|c|c|}
\hline
\multirow{2}{*}{min\_samples} & \multirow{2}{*}{Correction} & \multicolumn{13}{c|}{max\_depth} \\
\cline{3-15}
 & & 1 & 2 & 3 & 4 & 5 & 6 & 7 & 8 & 9 & 10 & 11 & 12 & 13 \\
\hline\hline
5 & no correction & 0.908 & 0.902 & 0.886 & 0.864 & 0.845 & 0.801 & 0.783 & 0.762 & 0.743 & 0.739 & 0.731 & 0.724 & 0.728 \\
5 & LOO & 0.912 & 0.901 & 0.885 & 0.872 & 0.877 & 0.873 & 0.871 & 0.870 & 0.869 & 0.875 & 0.869 & 0.867 & 0.872 \\
5 & Mallows & 0.908 & 0.902 & 0.884 & 0.874 & 0.875 & 0.869 & 0.873 & 0.868 & 0.869 & 0.868 & 0.870 & 0.867 & 0.872 \\
\hline
10 & no correction & 0.912 & 0.903 & 0.892 & 0.887 & 0.869 & 0.870 & 0.874 & 0.872 & 0.870 & 0.871 & 0.870 & 0.869 & 0.873 \\
10 & LOO & 0.912 & 0.903 & 0.890 & 0.884 & 0.876 & 0.880 & 0.886 & 0.886 & 0.889 & 0.879 & 0.886 & 0.885 & 0.888 \\
10 & Mallows & 0.911 & 0.905 & 0.890 & 0.886 & 0.886 & 0.888 & 0.881 & 0.885 & 0.888 & 0.882 & 0.888 & 0.883 & 0.885 \\
\hline
15 & no correction & 0.910 & 0.906 & 0.889 & 0.881 & 0.881 & 0.883 & 0.881 & 0.882 & 0.880 & 0.877 & 0.880 & 0.880 & 0.877 \\
15 & LOO & 0.910 & 0.906 & 0.889 & 0.887 & 0.887 & 0.879 & 0.889 & 0.885 & 0.884 & 0.886 & 0.883 & 0.890 & 0.885 \\
15 & Mallows & 0.910 & 0.901 & 0.894 & 0.886 & 0.883 & 0.884 & 0.886 & 0.887 & 0.885 & 0.885 & 0.889 & 0.886 & 0.889 \\
\hline
20 & no correction & 0.914 & 0.904 & 0.896 & 0.878 & 0.878 & 0.878 & 0.878 & 0.877 & 0.874 & 0.878 & 0.879 & 0.878 & 0.880 \\
20 & LOO & 0.913 & 0.901 & 0.886 & 0.885 & 0.883 & 0.884 & 0.881 & 0.884 & 0.880 & 0.883 & 0.882 & 0.886 & 0.886 \\
20 & Mallows & 0.912 & 0.902 & 0.885 & 0.882 & 0.889 & 0.886 & 0.888 & 0.883 & 0.885 & 0.885 & 0.882 & 0.887 & 0.884 \\
\hline
\end{tabular}
\end{table}


\subsection{Computational time}

\begin{figure}[!ht]
\centering
\includegraphics[scale=0.6]{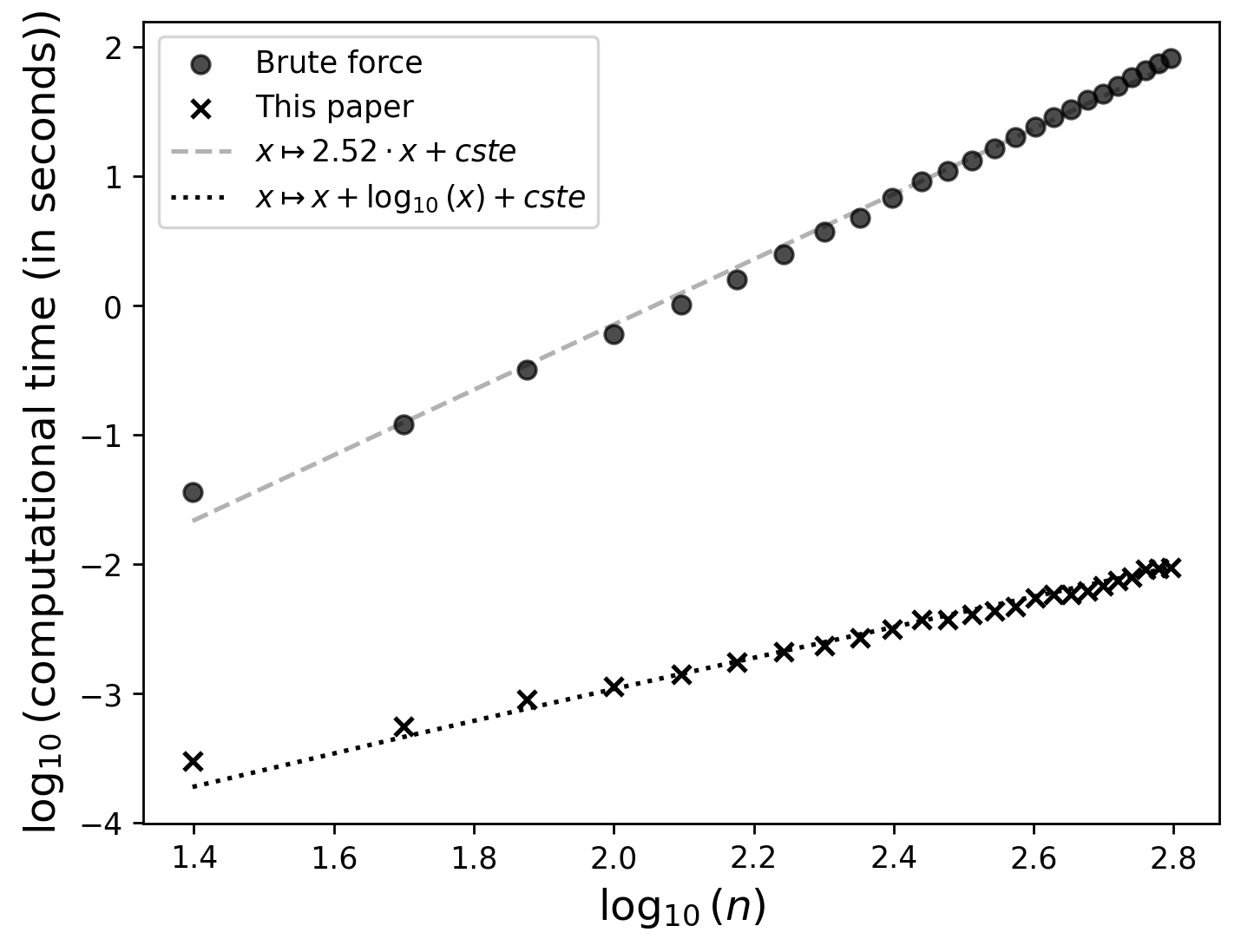}
\caption{Time to compute the CRPS of the prefix sequences $\textbf y^{(s)}=(y_1,\dots,y_s)$ for all $s \in [n]$ where $y_1,\dots, y_n$ are sampled independently from a standard normal distribution. For the brute force approach, we compute the CRPS using the {\it properscoring} Python package.}
\label{fig:compute_time_gaussian}
\end{figure}

In this section, we conduct experiments to assess the scalability of our method to learn regression trees minimizing the CRPS loss. First, we consider random vectors $(y_1,\dots, y_n)$ with i.i.d.\ entries sampled from a white noise Gaussian distribution and we compute the empirical entropy $ H_{\mathrm{CRPS}}(\textbf y^{(s)})$ associated to all prefix sequences $\textbf y ^{(s)}= (y_1,\dots,y_s)$ for all $s \in [n]$, namely
\[ H_{\mathrm{CRPS}}(\textbf y^{(s)}) := \frac1s \sum_{i=1}^s\mathrm{CRPS}(\widehat F^{(s)},y_i),\]
where $\widehat F^{(s)}$ is the empirical CDF of sequence $\textbf y ^{(s)}$. We compare with a method computing all CRPS independently using available Python packages. Figure~\ref{fig:compute_time_gaussian} demonstrates the computational time required for a node split during the training of a regression tree that minimizes the CRPS loss. When naively computing all CRPS values, the time complexity scales empirically as \( O(n^{2.55}) \). In contrast, the method proposed in this paper has a computational time complexity that scales as \( O(n \log n) \), as theoretically demonstrated.

\subsection{Comparing CRPS-RF and PMQRF with QRF}
\label{sec:expe_real_data}

In this section, we describe the numerical experiments conducted to evaluate the performance of our models. We selected six datasets from the UCI Machine Learning Repository to cover a variety of real-world applications. The total number of samples $n$ and features $d$ for each dataset are summarized in Table~\ref{tab:datasets}, with URLs pointing to the corresponding publicly available resources given in the Supplement.

\begin{table}[!ht]
\centering
\begin{tabular}{|c|c|c|c|c|c|c|}\cline{2-7}
\multicolumn{1}{c|}{} & {\bf Abalone} & {\bf GPU} & {\bf Turbine} & {\bf Cycle} & {\bf WineRed} & {\bf WineWhite} \\\hline
$n$ & 4177 & 241600 & 36733 & 9568 & 4898 & 1599 \\\hline
$d$ & 8 & 14 & 12 & 4 & 11 & 11 \\\hline
\end{tabular}
\caption{Meta-data for the datasets. $n$ refers to the total
number of data-points from which we create $300$ versions by independently drawing $1000$ data points randomly. $d$ refers to the feature dimension.}
\label{tab:datasets}
\end{table}
For each dataset, we randomly sampled 1,000 data points without replacement to generate the training sets, and 3,000 other samples used as the test set to compute the prediction error. We consider forests of PMQRTs or CRPS-RTs with 100 trees. Trees in the forests are trained by sampling uniformly at random without replacement $60\%$ of the training data. This procedure was repeated for 300 different training sets, each generated by sampling the data points randomly. We computed the mean and standard deviation of the CRPS on the test set across these 300 experiments, which are reported in Table~\ref{tab:res_real_data}. For each method, the CRPS was computed on the test set by querying quantiles with level $\mathcal Q_{0.05}:=\{0.05\times k, \quad k\in[19]\}$ and building the corresponding CDF. In the following, we present the different models considered in our experiments:
\begin{itemize}
\item[(CRPS-RF)] We train CRPS-RFs using the algorithm presented in Section~\ref{sec:CRPS}.
\item[(PMQRF)] The PMQRFs are trained using the list of quantile levels from $\mathcal Q_ {0.05}$. 
\item[(QRF)] To benchmark our models, we compared their performance with that of the Quantile Regression Forest (QRF) method from \citep{meinshausen2006quantile}.
\end{itemize}

An important aspect of the tree-based methods such as QRF, PMQRF, and CRPS-RF is that they rely on trees learning a single partition of the feature space for all quantile levels. This common partitioning has certain advantages, such as facilitating group conformal prediction (see Section~\ref{sec:group_conformel} for more details) and ensuring that the estimated quantiles are non-crossing. However, for some data distributions, the use of a single partition for all quantiles might lead to suboptimal results. To assess the cost of using a common partition for all quantile levels, we compared the aforementioned methods with:
\begin{itemize}
\item[(PQRTs-5)] We train 95 PQRTs where for each quantile level $\tau$ in $\mathcal Q_{0.05}$, $5$ PQRTs are trained specifically using the pinball loss $\ell_{\tau}$. At inference time, we query an estimate $\hat q_{\tau}$ by aggregating the forest of $5$ trees trained for the quantile level $\tau$. We then rearrange the entire set of esimated quantiles following the method proposed in~\cite{SQR}.
\item[(PMQRF-split)] An intermediate approach interpolating between the (PQRTs-5) and (PMQRF) methods, in which half of the trees ($50$ trees) are trained using the sum of pinball losses $\sum_{k=1}^{9}\ell_{0.05\times k}$ and the other half  ($50$ trees) with $\sum_{k=10}^{19}\ell_{0.05\times k}$.
\item[(SKLearn)] A quantile regression model from the Scikit-learn library, where an independent linear model is trained for each quantile level. This allows for more flexible partitioning of the feature space, though it comes at the cost of losing the benefits of a shared partition. 
\end{itemize}

Table~\ref{tab:res_real_data} presents the comparative results obtained across all datasets. Overall, the findings reveal several consistent patterns regarding the relative performance and robustness of the evaluated methods.

\paragraph{Limitations of parametric models.}
The SKLearn model achieves the best performance only on the \textit{Turbine} dataset, where it significantly outperforms all other methods. However, it ranks as the worst model on every other dataset, often by a large margin. This behavior highlights the strong dependence of SKLearn on its underlying linearity assumption: when the true conditional quantiles can be well approximated by a linear model, SKLearn performs remarkably well, but its accuracy deteriorates substantially when this assumption is violated.

\paragraph{CRPS-RF and PMQRF as robust, assumption-free methods.}
Across datasets, the CRPS-RF consistently provides the most accurate or at least competitive results among all tree-based approaches. Similarly, the PMQRF demonstrates strong and stable performance. These two methods share an important property: they are assumption-free and trained using interval-based scores (namely, the WIS or the CRPS; see~\cite[Section 3.2]{WIS}). This formulation allows them to model a wide variety of conditional distributions without relying on restrictive parametric forms, which explains their robustness across heterogeneous datasets.

\paragraph{Behavior of PQRTs-5 and related variants.}
The PQRTs-5 method underperforms compared to both PMQRF and CRPS-RF, reflecting its limited robustness when the loss function focuses on a restricted set of quantiles. This behavior suggests that extreme quantiles are more difficult to estimate accurately in isolation, while jointly learning several quantiles tends to stabilize training and yield more reliable predictions. The intermediate variant—where half of the trees in the forest are trained on lower quantiles ($\sum_{k=1}^{9}\ell_{0.05\times k}$) and the other half on upper quantiles ($\sum_{k=10}^{19}\ell_{0.05\times k}$)—does not lead to better results than PMQRF. Nevertheless, we believe that this strategy could be beneficial for datasets where the behaviors of the lower and upper tails differ substantially.

\paragraph{Assumption-based versus assumption-free models.}
It is worth emphasizing that both SKLearn and QRF can be viewed as assumption-based models. SKLearn assumes linearity in the quantile relationship, while QRF relies on standard regression trees that implicitly assume a Gaussian conditional distribution through the use of variance as the impurity measure. This variance corresponds to the generalized entropy (negative log-likelihood) of a Gaussian model, which effectively optimizes a logarithmic score.
In contrast, PMQRF and CRPS-RF are trained without such distributional assumptions, relying instead on interval-based scoring rules. These scores are known to be more robust than logarithmic scores \citep{WIS}, particularly when the true data distribution deviates from Gaussianity. Two main conclusions emerge from these experiments:
\begin{itemize}
\item Assumption-based models (e.g., SKLearn, QRF) can perform extremely well when their underlying assumptions hold—such as linearity or Gaussianity—but their performance deteriorates markedly when those assumptions are violated.
\item PMQRF and CRPS-RF may not always be optimal for specific data distributions, yet they tend to yield competitive and stable results across diverse datasets. Their robustness stems from both their assumption-free nature and their reliance on interval-based scoring rules.
\end{itemize}

\begin{table}[!ht]
\centering
\begin{tabular}{|c|cccccc|}\hline
{\bf Method} & Abalone & GPU & Turbine & Cycle & WineRed & WineWhite \\\hline\hline
QRF & \makecell{1.93 \\ (0.02)} & \makecell{104.35 \\ (5.02)} & \makecell{1.31 \\ (0.02)} & \makecell{4.11 \\ (0.04)} & \makecell{\bf 0.53 \\ \bf (0.01)} & \makecell{\bf 0.63 \\ \bf (0.00)} \\\hline
PQRTs-5 & \makecell{2.00 \\ (0.03)} & \makecell{155.27 \\ (5.41)} & \makecell{1.88 \\ (0.05)} & \makecell{4.43 \\ (0.05)} & \makecell{0.57 \\ (0.00)} & \makecell{0.67 \\ (0.01)} \\\hline
PMQRF-split & \makecell{\bf 1.86 \\ \bf (0.02)} & \makecell{103.99 \\ (5.28)} & \makecell{1.13 \\ (0.02)} & \makecell{\bf 3.88 \\ \bf (0.04)} & \makecell{\bf 0.53 \\ \bf (0.01)} & \makecell{\bf 0.63 \\ \bf (0.00)} \\\hline
PMQRF & \makecell{\bf 1.86 \\ \bf (0.02)} & \makecell{\bf 97.15 \\ \bf (5.01)} & \makecell{1.12 \\ (0.02)} & \makecell{\bf 3.88 \\ \bf (0.04)} & \makecell{\bf 0.52 \\ \bf (0.01)} & \makecell{\bf 0.63 \\ \bf (0.01)} \\\hline
CRPS-RF & \makecell{\bf 1.88 \\ \bf (0.02)} & \makecell{\bf 94.83 \\ \bf (5.35)} & \makecell{1.11 \\ (0.02)} & \makecell{\bf 3.87 \\ \bf (0.04)} & \makecell{0.59 \\ (0.01)} & \makecell{0.68 \\ (0.01)} \\\hline
SKLearn & \makecell{2.00 \\ (0.02)} & \makecell{182.47 \\ (7.60)} & \makecell{\bf 0.58 \\ \bf (0.01)} & \makecell{4.51 \\ (0.05)} & \makecell{0.63 \\ (0.01)} & \makecell{0.74 \\ (0.01)} \\\hline
\end{tabular}
\caption{For each method and dataset, the first value represents the mean of the CRPS on the test sets, while the second value in parentheses indicates its standard deviation across the 300 simulations. In bold, we highlight the best-performing method (i.e., the one with the lowest mean CRPS error), as well as all other methods whose mean CRPS error is below the mean plus one standard deviation of the best-performing method.}
\label{tab:res_real_data}
\end{table}



\subsection{Adaptive group conformal prediction with CRPS random trees}
\label{sec:group_conformel}
In this section, we show how CRPS-RTs can be used to achieve group conformal prediction. Inspired by the simulated dataset from \cite{sesia2020comparison}, we consider data $\{(\textbf x_i,y_i)\}_{i=1}^n$ where 
 \begin{equation}\label{eq:data}\forall i \in [n],\quad y_i = f(\textbf x_i^{\top}\beta) + \epsilon_i \sqrt{1+\big(\textbf x_i^{\top} \beta\big)^2},\end{equation}
 with $\beta = [1\,, \,1] \in \mathbb R^2$, $\textbf x_i \sim \mathrm{Unif}([0,1]^2)$ are independent vectors, and $(\epsilon_i)_{i\in [n]}$ are independent standard normal random variables.

We train a single CRPS-RT with a training set of size $n=2000$. We define four regions corresponding to the partition of the feature space obtained from the CRPS-RT at depth 2. We then evaluated different conformalized predictive inference methods, based on CRPS-RF or QRF with 50 trees, with both marginal and group-based conformalization considering the four previously defined groups. Our goal is to assess how well these methods achieve the desired coverage while maintaining narrow prediction intervals across the different regions.

We use the split conformal framework to conformalize the different methods. To do so, we consider a calibration set with $1000$ data points and a test set of $5000$ points. We conformalize the CRPS-RF method by considering the family of nested sets presented in \cite{chernozhukov2021distributional}, namely of the form 
\[[\hat q_t(\textbf x),\hat q_{1-t}(\textbf x)],\]
where $\hat q_t(\textbf x)$ denotes the empirical quantile of level $t$ obtained from our CRPS-RT by dropping the feature vector $\textbf x \in [0,1]^2$ from the root in the CRPS-RT and by taking the empirical quantile of level $t$ from the data points in the leaf where we end up. We compare our method with a similar split conformal method obtained by training QRF and conformalizing using the CQR method from \cite{romano2019conformalized}. In both case, we target a marginal coverage of $90\%=1-\alpha$.

\begin{figure}[!ht]
\centering
\includegraphics[scale=0.44]{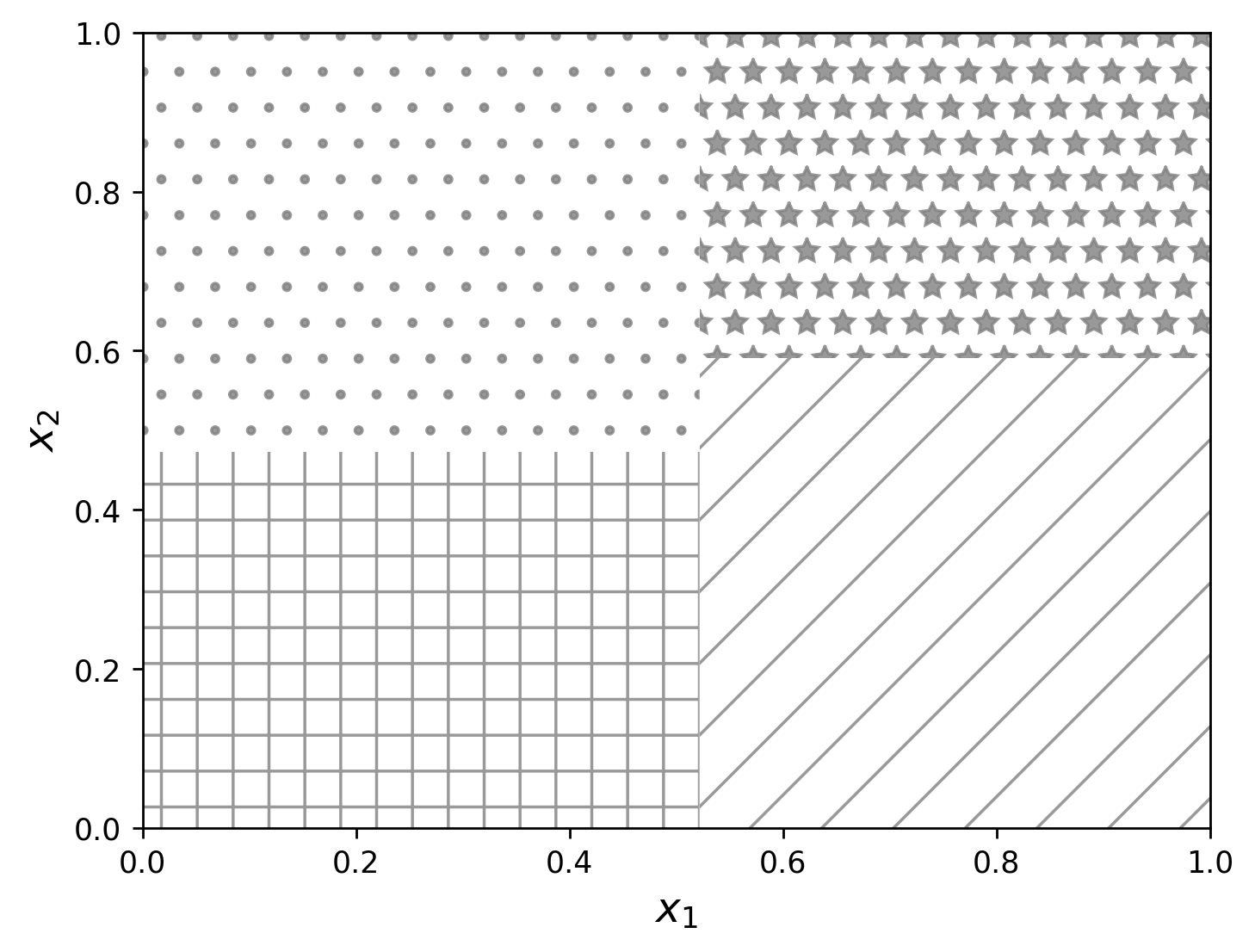}\\
\begin{tabular}{c|cccc|c}
{\bf Coverage} & Dot & Star & Grid & Hachure & Marginal \\\hline
QRF-CQR & 0.94 & 0.84 & 0.82 & 0.93 & 0.89 \\
Group QRF-CQR & 0.88 & 0.89 & 0.91 & 0.90 & 0.90 \\
CRPS-RF-distri & 0.88 & 0.87 & 0.93 & 0.89 & 0.89 \\
Group CRPS-RF-distri & 0.89 & 0.90 & 0.90 & 0.89 & 0.89 \\~\\
{\bf Mean width} &  Dot & Star & Grid & Hachure & Marginal \\\hline
QRF-CQR & 4.55 & 4.58 & 5.25 & 4.56 & 4.71 \\
Group QRF-CQR & 4.06 & 4.94  & 6.35 & 4.36 & 4.83 \\
CRPS-RF-distri & 3.72 & 4.62 & 6.39 & 4.67 & 4.74\\
Group CRPS-RF-distri & 3.85 & 4.74 & 5.93 & 4.29 & 4.70 \\
\end{tabular}
\caption{We train a single CRPS-RT and identify four distinct regions in the feature space, based on the learned partition induced by the CRPS-RT at depth 2. We evaluate the performance of conformalized predictive inference methods, including CRPS-RF (with both marginal and group-based conformalization) and QRF-CQR. For each method, we report the coverage and mean prediction interval width across the four regions.}
\label{fig:groupcov}
\end{figure}

Figure~\ref{fig:groupcov} illustrates the impact of different conformalization strategies on coverage and mean prediction interval width across the four regions identified by the CRPS-RT partitioning. Notably, the QRF-CQR method without group conformalization results in highly unbalanced coverage, failing to maintain the desired coverage level uniformly across the regions. This suggests that standard marginal conformalization struggles to adapt to the heterogeneity present in the learned feature space partition. In contrast, the CRPS-RF method with a distributional nested set and marginal calibration naturally provides better balance in coverage across the four regions. This improved calibration highlights the advantage of leveraging CRPS-based probabilistic forests for predictive inference.

Among all the methods considered, the group-conformalized CRPS-RF approach emerges as the most effective. It not only achieves the target coverage across all four regions but also maintains the narrowest marginal mean prediction interval width. This demonstrates that incorporating group-based conformalization within CRPS-RF leads to both improved calibration and sharper predictive intervals, making it the most reliable approach in this setting.

\definecolor{MyGreen}{RGB}{180, 220, 180}
\definecolor{MyRed}{RGB}{250, 180, 180}
\definecolor{MyBlue}{RGB}{180, 180, 250}
\definecolor{MyYellow}{RGB}{250, 220, 150}

\section{Conclusion}

We have shown that it is possible to construct tree-based algorithms that minimize either the sum of $M$ pinball losses or the CRPS, with computational complexity for one node split of, respectively, $O(Mdn\log n)$ and $O(dn\log n)$. In this setting, we also established that improved estimates of the population entropies can be obtained at no additional computational cost, by using LOO or Mallows-type corrections, leading to better generalization performance for the resulting trees and random forests. Empirically, PMQRFs and CRPS-RFs were shown to perform similarly to, or better than, existing approaches such as QRF.

These trees are particularly useful in the context of conformal prediction, since informative conformal intervals can be constructed from at least pairs of quantiles \citep{romano2019conformalized}, or even from conditional CDFs \citep{chernozhukov2021distributional}. Moreover, shallow trees produced by our algorithms can be used to define an interpretable partition of the feature space, thereby yielding natural groups on which group conformal prediction methods can be applied.

\noindent \textbf{Disclosure statement}: The authors report there are no competing interests to declare.

\bibliographystyle{chicago}
\bibliography{sample}

\clearpage

\appendix

\bigskip
\begin{center}
{\large\bf APPENDIX}
\end{center}

\section{Background on Data Structures and Computational Tools}
\label{apdx:data_structures}

In this section, we provide a brief, self-contained introduction to the key data structures and computational tools used in our distributional regression tree algorithms.

We also provide concrete examples illustrating how our algorithms for learning PMQRT and CRPS-RT update their respective data structures — min–max heaps and Fenwick trees — throughout the iterations. We present the updates for selected steps to highlight the key mechanisms, while the full, step-by-step versions of these examples are available in the documentation of our package (section {\it Algorithms unraveled}) for interested readers.

\subsection{Heaps}

\paragraph{Presentation of the data structures.}

A heap is a specialized tree-based data structure that allows efficient access to extreme values in a dataset. In a max-heap, each parent node is greater than or equal to its children, ensuring that the maximum element is always at the root. Conversely, in a min-heap, each parent node is less than or equal to its children, so the minimum element is at the root. Both max-heaps and min-heaps are useful in statistical algorithms when we need to track the largest or smallest values efficiently, such as maintaining quantile thresholds or top-k observations. A min-max heap generalizes these ideas to provide fast access to both the minimum and maximum elements simultaneously. In a min-max heap, nodes at even levels (with the root at level 0) are smaller than all their descendants, while nodes at odd levels are larger than all their descendants. This guarantees that the minimum element is at the root and the maximum element is at one of its children. A concrete example of min-max heap is presented in Figure~\ref{fig:minmaxheap}. 

\begin{figure}[h!]
\centering
\begin{tikzpicture}[level distance=1.2cm,
  level 1/.style={sibling distance=3cm},
  level 2/.style={sibling distance=1.5cm}]

\node[circle, draw] {1} 
  child {node[circle, draw] {10} 
    child {node[circle, draw] {3}} 
    child {node[circle, draw] {5}}
  }
  child {node[circle, draw] {8} 
    child {node[circle, draw] {4}} 
    child {node[circle, draw] {6}}
  };
\end{tikzpicture}
\caption{Example of a min-max heap: even levels (0,2,...) store minimums relative to descendants, odd levels (1,3,...) store maximums.}
\label{fig:minmaxheap}
\end{figure}

\paragraph{Practical role in this work.}

\begin{figure}[htbp]
    \centering
    \begin{subfigure}{0.45\textwidth}
        \includegraphics[width=\linewidth,page=1,trim=1.2cm  2cm 1.9cm 1cm,clip]{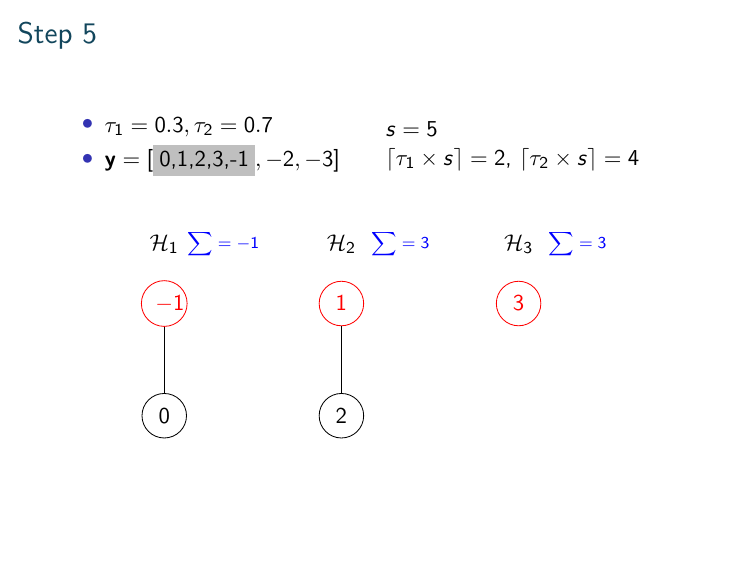}
        \caption{Step 5}
    \end{subfigure}
    \begin{subfigure}{0.45\textwidth}
        \includegraphics[width=\linewidth,page=2,trim=1.2cm  2cm 1.9cm 1cm,clip]{heaps.pdf}
        \caption{Step 6}
    \end{subfigure}
    \caption{Illustration of an update of the min-max heaps at a specific iteration of the algorithm to perform a node split in a PMQRT. We consider the sum of pinball losses with quantile levels $\tau_1=0.3$ and $\tau_2=0.7$. At step 6, we need to insert $y_6=-2$ in the min-max heaps. $\mathcal H_1$ must contain $\ceil{\tau_1 \times 6}=2$ elements and it already contains two elements. Since $-2$ is smaller than the current maximum element of $\mathcal H_1$ (namely $0$), we pop out $0$ and insert $-2$ in $\mathcal H_1$. Now, we need to insert $0$ in one of the remaining heaps $\mathcal H_2,\mathcal H_ 3$.  $\mathcal H_2$ must contain $\ceil{\tau_2 \times 6}-\ceil{\tau_1 \times 6}=5-2=3$ elements, and it currently contains $2$ elements. Since $0$ is smaller than the current maximum value in $\mathcal H_2$ (namely $2$), we insert $0$ in $\mathcal H_2$ and the update is completed.}
    \label{fig:example_PMQRT}
\end{figure}

In our work, min–max heaps are used to easily maintain and update order statistics within each node during split evaluation, allowing us to efficiently perform a node splits for the generalized entropy associated to the WIS loss.  Figure~\ref{fig:example_PMQRT} gives a concrete example on the way the min-max heaps are updated at a given iteration to perform a node split in a PMQRT. We consider as loss function the sum of two pinball losses with quantile levels respectively $0.3$ and $0.7$. We assume that we need to identify the best split for a given feature $j$ for which $x_{1,j}\leq \dots\leq x_{7,j}$ with corresponding observations $\textbf y= [0,1,2,3,-1,-2,-3]$. In Figure~\ref{fig:example_PMQRT}, we illustrate the evolution of the three min-max heaps $\mathcal H_1$, $\mathcal H_2$ and $\mathcal H_3$ from iteration $5$ to $6$ to compute the entropies associated to the prefix sequence $\textbf y^{(6)}$. Next to each heap, the blue value represents the sum of all elements within the corresponding heap.

\subsection{Prefix and Suffix Sequences}

\paragraph{Presentation of the data structures.}

A prefix sequence is the subsequence starting from the first element up to a given position. Conversely, a suffix sequence starts from a given position and continues to the last element. Figure~\ref{fig:prefix_seq} illustrates the prefix and suffix sequences on a concrete example.

\begin{figure}[h!]
\centering
\begin{tikzpicture}[scale=1, every node/.style={scale=0.9}]
\node at (0,0) {Original sequence: $[3, 1, 4, 2]$};
\node at (0,-0.6) {Prefix sequences: $[3]$, $[3,1]$, $[3,1,4]$, $[3,1,4,2]$};
\node at (0,-1.2) {Suffix sequences: $[3,1,4,2]$, $[1,4,2]$, $[4,2]$, $[2]$};
\end{tikzpicture}
\caption{Prefix sequences start from the first element to a given position. Suffix sequences start at a given position to the end.}
\label{fig:prefix_seq}
\end{figure}

\paragraph{Practical role in this work.}

Prefix and suffix sequences are convenient notations in our algorithms since, when evaluating candidate splits, we must compute the entropy for both the prefix sequence (samples that fall to the left of the split) and the suffix sequence (samples that fall to the right). More precisely, in a given leaf containing data points $\{(x_i, y_i)\}_{i=1}^n$, for which we want to compute the information gains associated with all possible splits on feature $j$, we assume without loss of generality that
\[
x_{1,j} \leq x_{2,j} \leq \dots \leq x_{n,j}.
\]
We then need to evaluate the entropies of the prefix sequence $(y_1, \dots, y_s)$ and the suffix sequence $(y_{s+1}, \dots, y_n)$ for all $s \in [n]$ in order to compute the information gain corresponding to feature $j$ and split position $s$ in the current leaf.

\subsection{Weight-Balanced Binary Trees}

\paragraph{Presentation of the data structures.}

A weight-balanced binary tree is a type of binary search tree in which the sizes of left and right subtrees are maintained roughly equal to ensure balanced height. This guarantees that insertion, deletion, and query operations can be performed in logarithmic time. 

\paragraph{Practical role in this work.}

In our work, weight-balanced trees are used in the construction of CRPS-RTs. When computing the information gains for a feature $j$ within a given leaf, we iteratively evaluate the gains for all possible splits $s \in [0, n]$ by computing the entropies of the corresponding prefix and suffix sequences, $(y_1, \dots, y_s)$ and $(y_{s+1}, \dots, y_n)$ (where we consider without loss of generality that $x_{1,j} \leq x_{2,j} \leq \dots \leq x_{n,j}$).

To perform these iterations efficiently, we need to determine the insertion rank of the element $y_{s+1}$ within the sorted version of the list $\mathbf{y}^{(s)} = (y_1, \dots, y_s)$. This rank is crucial for efficiently updating the Fenwick tree, which we use to compute the information gains associated with all possible splits for a given feature $j$.

A naive approach would involve searching for the insertion rank of $y_{s+1}$ in the sorted version of $\mathbf{y}^{(s)}$ at each iteration, resulting in prohibitive computational cost. To mitigate this, given a feature $j$, we first iteratively insert the elements $y_1, y_2, \dots, y_n$ into a Weight-Balanced Binary Tree. This allows us to obtain, in $\mathcal{O}(n \log n)$ time, the rank of insertion of $y_{s+1}$ in the sorted version of $\mathbf{y}^{(s)}$ for any $s \in [n]$.

These precomputed insertion ranks, denoted by $\{r_s\}_{s=1}^n$, are then used to update the Fenwick tree efficiently during the computation of information gains. In the next subsection, we provide a concrete example illustrating how the Fenwick tree is updated using these precomputed ranks.

Figure~\ref{fig:WBT} shows how a Weight-Balanced Binary Tree can be used to determine the rank of a new element in a sorted list.

\begin{figure}[!h!]
\centering
\begin{tikzpicture}[level distance=1.2cm,
  level 1/.style={sibling distance=3cm},
  level 2/.style={sibling distance=1.5cm}]
\node[circle, draw] {10}
  child {node[circle, draw] {5}
    child {node[circle, draw] {2}}
    child {node[circle, draw] {7}}
  }
  child {node[circle, draw] {15}
    child {node[circle, draw] {12}}
    child {node[circle, draw] {18}}
  };

\node[align=left, text width=9cm] at (8,-0.5) {
Example: Determine insertion rank of 6 in the sorted list \([2,5,7,10,12,15,18]\) \\
- Start at root (10): $6 < 10$, go left, current rank = 0 \\
- Node 5: $6 > 5$, go right, add size of left subtree + 1 → rank = 1 + 1 = 2 \\
- Node 7: $6 < 7$, go left and stop, final rank = 2
};

\end{tikzpicture}
\caption{Using a weight-balanced binary tree to determine the rank of an element in a sorted list. The rank is computed by counting nodes in left subtrees along the search path.}
\label{fig:WBT}
\end{figure}

\subsection{Fenwick Trees}

\paragraph{Presentation of the data structures.}

A Fenwick tree, also called a binary indexed tree, is a data structure designed to efficiently compute and update prefix sums over a sequence of numbers. Unlike a simple prefix sum array, a Fenwick tree allows efficient dynamic updates: if an element in the sequence changes, all relevant prefix sums can be updated in logarithmic time, without recomputing sums for the entire array.

Formally, given a sequence $(x_1, x_2, \dots, x_n)$, the Fenwick tree stores partial sums that allow two operations in $\mathcal O(\log n)$ time:

\begin{enumerate}
\item Query: compute the sum of the first $i$ elements, $\sum_{k=1}^i x_k$.
\item Update: add a value to a particular element $x_i$ and update all affected prefix sums.
\end{enumerate}
We refer to Figure~\ref{fig:fenwick_tree} for a visualization of the structure of a Fenwick tree.

\paragraph{Practical role in this work.}

\begin{figure}[htbp]
    \centering
    \begin{subfigure}{0.45\textwidth}
        \includegraphics[width=\linewidth,page=1,trim=1cm  0.5cm 0cm 0cm,clip]{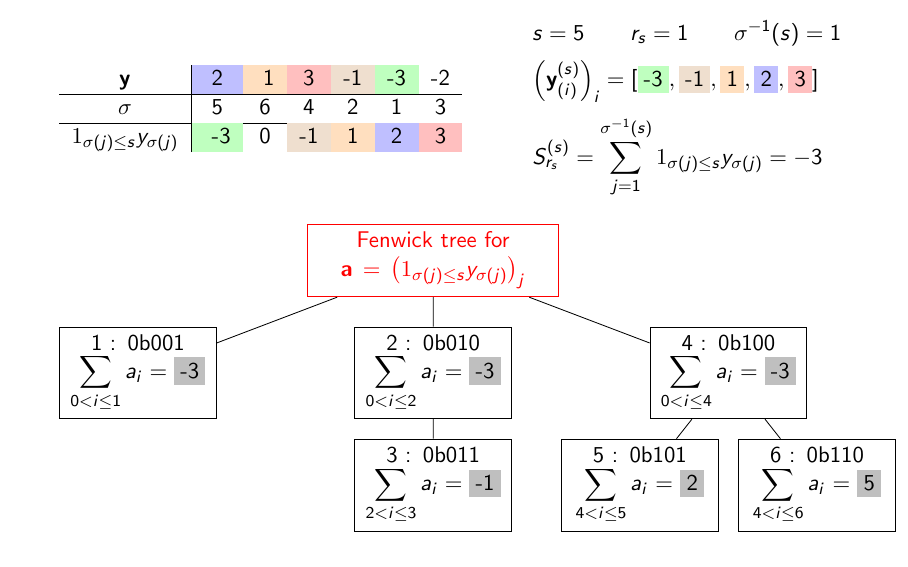}
        \caption{Step 5}
    \end{subfigure}
    \begin{subfigure}{0.45\textwidth}
        \includegraphics[width=\linewidth,page=2,trim=1cm  0.5cm 0cm 0cm,clip]{fenwick.pdf}
        \caption{Step 6}
    \end{subfigure}
    \caption{Illustration of an update of the Fenwick tree at a specific iteration of the algorithm to perform a node split in a CRPS-RT. We first insert $y_s=-2$ in the Fenwick tree at the position $\sigma^{-1}(s)$, with $s=6$. Then, we know that querying the prefix sum up to index $\sigma^{-1}(s)$ will give us the partial sum $S^{(s)}_{r_s}$, needed to update the entropy.}    \label{fig:example_CRPSRT}
\end{figure}

We provide a concrete example to illustrate how our algorithm to learn CRPS-RT update the Fenwick tree along the iterations. Figure~\ref{fig:example_CRPSRT} illustrates the updates of the Fenwick tree to perform a node split in a CRPS-RT. Similarly to Figure~\ref{fig:example_PMQRT}, we assume that we need to identify the best split for a given feature $j$ for which $x_{1,j}\leq \dots\leq x_{6,j}$ with corresponding observations $\textbf y= [2,1,3,-1,-3,-2]$.

\section{Study of different aggregation strategies}
\label{apdx:study_agg}

We evaluate the performance of the two aggregation methods described in the introduction of Section~\ref{sec:expes}, namely distributional bagging (DB) and quantile bagging (QB). For each dataset introduced in Section~\ref{sec:expes}, we randomly subsample 1,000 points to train a CRPS-RF consisting of 100 trees. The maximum depth was fixed at 10, the $min\_samples\_split$ parameter was set to 30, and LOO was used to compute the information gains. We then query the $5\%$ and $95\%$ quantiles on a test set of size 1,000 and compute both the marginal width of the predictive intervals $[\hat q_{0.05},\hat q_{0.95}]$ and the corresponding marginal coverage. This procedure is repeated over 100 different subsamples of the original dataset, allowing us to report the mean and standard deviation of the marginal width and coverage across these scenarios. Table~\ref{tab:aggregation} summarizes the results, showing that the QB aggregation scheme achieves a marginal coverage close to the nominal $90\%$, while the DB scheme tends to be more conservative, resulting in wider predictive intervals and higher coverage than expected. For this reason, the QB aggregation scheme was preferred in the experiments presented in our paper.

\begin{table}[h!]
\centering
\caption{Mean and standard deviation of marginal widths and coverage when using either the DB or QB aggregation strategy.}
\label{tab:aggregation}
\begin{tabular}{|c|cc|cc|}
\hline
\multirow{2}{*}{Dataset} & \multicolumn{2}{c|}{Width} & \multicolumn{2}{c|}{Coverage} \\
 & DB & QB & DB & QB \\
\hline
Abalone & 7.88 $\pm$ 0.21 & 7.43 $\pm$ 0.14 & 0.95 $\pm$ 0.01 & 0.90 $\pm$ 0.01 \\
GPU & 578.93 $\pm$ 18.26 & 528.59 $\pm$ 20.47 & 0.92 $\pm$ 0.01 & 0.91 $\pm$ 0.01 \\
Turbine & 7.73 $\pm$ 0.12 & 6.44 $\pm$ 0.12 & 0.96 $\pm$ 0.01 & 0.93 $\pm$ 0.01 \\
Cycle & 15.77 $\pm$ 0.26 & 14.51 $\pm$ 0.15 & 0.92 $\pm$ 0.01 & 0.90 $\pm$ 0.01 \\
WhiteWine & 2.49 $\pm$ 0.11 & 2.51 $\pm$ 0.08 & 0.97 $\pm$ 0.01 & 0.92 $\pm$ 0.01 \\
\hline
\end{tabular}
\end{table}

\section{Package description}
\label{sec:package}

Table \ref{tab:config-options} summarizes the main configuration options available in the {\it DisTreebution} package. The parameters are organized into three conceptual layers reflecting the structure of the modeling framework. The first layer defines the model type, allowing users to choose between CRPS regression trees (CRPS-RTs) and pinball-loss-based quantile regression trees (PMQRTs). The second layer specifies the tree training settings, which control how individual trees are constructed and aggregated within the ensemble. These include standard hyperparameters such as tree depth and minimum samples per split, as well as more advanced options like the use of a leave-one-out estimate of the information gain for data-driven split selection, and the strategy for aggregating leaf predictions. The final layer relates to the conformalization settings, which govern the calibration of predictive intervals. Users can choose between a distributional or conformalized quantile regression (CQR) approach, optionally applying conformalization at the group level. Together, these parameters offer a flexible yet principled way to configure {\it DisTreebution} for a range of uncertainty quantification tasks.

\paragraph{Quantile-level configuration for PMQRTs.}
When using PMQRTs, the dictionary $treeID2quantiles\_train$ specifies the set of quantile levels on which each tree is trained. More concretely, setting $\left\{i:[\beta,1-\beta]\, \text{for} \, i \in [nTrees]\right\}$ trains all trees in the forest using the combined pinball loss $\ell_{\beta}+\ell_{1-\beta}$. Alternatively, defining $\left\{i:[\beta], \text{for} , i \in [nTrees/2]\right\}\cup\left\{i:[1-\beta]\, \text{for} \, i \in \{nTrees/2,\dots,nTrees\}\right\}$ results in half of the trees being optimized with $\ell_{\beta}$ and the other half with $\ell_{1-\beta}$. By default, each tree can be queried only for the quantile levels it has been trained on, although this behavior can be manually adjusted by the user. We refer readers to the package documentation for additional configuration options. This design provides flexibility, allowing PMQRFs to distribute modeling effort across different quantile levels. 

\begin{table}[h!]
\centering
\small
\caption{Configuration options for the {\it DisTreebution} package}
\label{tab:config-options}
\begin{tabular}{|c|C{5cm}|C{9cm}|}
\hline
\textbf{Layer} & \textbf{Parameter}  & \textbf{Notes} \\ \hline
\multirow{2}{*}{\raisebox{0.4\normalbaselineskip}[0pt][0pt]{\rotatebox[origin=c]{90}{\makecell{\textbf{I. Model}\\\textbf{Type}}}}}
 & \makecell{type\_tree |\\  $\in \{CRPS, PMQRT,$ \\$RT\}$} & \makecell{CRPS: tree trained to minimize the CRPS. \\ PMQRT: tree trained to minimize a sum of pinball \\losses. \\ RT: standard regression tree.}\\ \hline
\multirow{6}{*}{\raisebox{-2\normalbaselineskip}[0pt][0pt]{\rotatebox[origin=c]{90}{\makecell{\textbf{II. Tree Training Settings}}}}}
 & nTrees | Int & Number of trees in the ensemble. \\ 
 & IG\_biais\_correction | $\in \{None, LOO, Mallows\}$ & Use leave-one-out or Mallows-style unbiased estimate of information gain for split selection. \\
 & type\_aggregation\_trees | $\in\{\text{QB,DB}\}$ & Method to aggregate predictions in leaves. \\
 & treeID2quantiles\_train | Dict of Lists & Specifies which quantile levels each tree is optimized for (for PMQRTs). \\
 & max\_depth | Int & Maximum depth of each tree. \\
 & min\_samples\_split | Int & Minimum number of samples required to attempt a split. \\
  & max\_depth\_group | Int & Depth up to which trees share the same partition of the feature space. This is useful only when one plans to use group conformal prediction. \\ \hline
\multirow{3}{*}{\rotatebox[origin=c]{90}{\makecell{\textbf{III. Conformalization}\\ \textbf{Settings}}}}
 & nested\_set | $\in \{$Distributional , CQR$\}$ & Type of nested-set used if conformalization is active. distributional: nested sets are $[\hat q_{t}, \hat q_ {1-t}]$; CQR: nested sets are of the form $[\hat q_{\beta}-t,\hat q_{1-\beta}+t]$ where $\beta$ is the nominal quantile. \\
 & nominal\_quantile & Nominal quantile level $\beta$ used for CQR, i.e. nested sets are of the form $[\hat q_{\beta}-t,\hat q_{1-\beta}+t]$ \\
 & group\_coverage | Bool  & Apply conformalization at group level. \\ \hline
\end{tabular}
\end{table}

\clearpage

\bigskip
\begin{center}
{\large\bf SUPPLEMENTARY MATERIAL}
\end{center}

\setcounter{section}{0} 
\renewcommand{\thesection}{\arabic{section}} 

\section{List of notations}

{\renewcommand{\arraystretch}{1.5}
\begin{table}[H]
\centering
\begin{tabular}{|c|L{14cm}|}\hline
$\textbf y^{(s)}$ & Prefix sequence $(y_1,\dots,y_s)$ of the list $(y_1,\dots,y_n)$\\\hline
$\textbf y^{(s)}_{(i)}$ & $i$-th larger element of the list $\textbf y^{(s)}$\\\hline
$r_s$ & The position to insert the element $y_s$ into the list $(y^{(s-1)}_{(1)},\dots, y^{(s-1)}_{(s-1)})$ in order to maintain its non decreasing order\\\hline
$S^{(s)}$ & $\sum_ {i=1}^s y^{(s)}_{(i)}=\sum_ {i=1}^s y_i$\\\hline
$S^{(s)}_r$ & $\sum_ {i=1}^r y^{(s)}_{(i)}$\\\hline
$W^{(s)}$ & $\sum_ {i=1}^s iy^{(s)}_{(i)}$\\\hline
$\sigma$ & Permutation sorting the list $\textbf y$ in increasing order, i.e. $y_{\sigma(1)} \leq \dots \leq y_{\sigma(n)} $ \\ \hline
$.get(a,b)$ & Operator acting on a Fenwick tree that outputs the cumulative sum of the elements from position $a$ to position $b$ (both included) in the corresponding tree.\\\hline
\end{tabular}
\label{table:notations}
\end{table}}

\section{Algorithms pseudo-codes for PMQRTs and CRPS-RTs}
\label{apdx:algos}
\subsection{Computations of the entropies for all splits for multiple quantiles}
\label{apdx:regression_tree_multi_quantiles}

We consider $M$ quantile levels: $\tau_1<\dots <\tau_M$. In this section, we provide the detailed algorithm to train regression trees to minimize the empirical entropy associated with a sum of the pinball losses with quantile levels $\tau_1,\dots,\tau_M$. By convention, querying the minimum or the maximum of an empty heap returns a $\text{None}$ value. Moreover, each min-max heap data structure keeps track of the total sum of the current elements within the heap.

Algorithm~\ref{alg:pmqrt} describes the procedure to compute the information gain induced by all the possible splits at a given node.

\begin{algorithm}
\caption{PMQRT: Returns the feature index and the threshold value giving the split of the data with the largest information gain.}\label{alg:pmqrt}
\begin{algorithmic}[1]
 \State {\bf Input}: Data in a given node of the leaf $\{(\textbf x_i,y_i)\}_{i=1}^n$
\State $score^{\star}, split^{\star}, feature^{\star}\gets + \infty, None, None$
\For{$j=1$ to $d$}
    \State Let $x_{i_1 ,j} \geq \dots \geq  x_{i_n,j}$ be the decreasing sorted $(x_{i,j} )_{i\in [n]}$ and set $\textbf y:=\big( y_{i_1},\dots,y_{i_{n}}\big).$
    \State $\ell_{\mathcal E}^{\downarrow} \gets $  GetEntropiesPMQRT($\textbf y)$ \quad {\color{blue}$/\star$\text{ cf. Algorithm~\ref{algo:GetEntropiesPMQRT}}} \vspace{0.2cm}
    \State Let $x_{i_1 ,k} \leq \dots \leq  x_{i_{n},k}$ be the increasing sorted $(x_{i,k} )_{i \in [n]}$ and set $\textbf y:=\big( y_{i_1},\dots,y_{i_{n}}\big).$
    \State $\sigma \gets \mathrm{ArgSort}(\textbf y) $\;
    \State $\ell_{\mathcal E}^{\uparrow} \gets $ GetEntropiesPMQRT$(\textbf y)$ \quad {\color{blue}$/\star$\text{ cf. Algorithm~\ref{algo:GetEntropiesPMQRT}}} \vspace{0.2cm}
    
    \State $w^{\star} , score \gets \text{ArgMin}\big(w\times\ell_{\mathcal E}^{\uparrow}[w]+(n-w)\times\text{Flip}(\ell_{\mathcal E}^{\downarrow})[w], \quad w \in \{0,\dots, n\}\big)$
    \If{$score < score^{\star}$}
        \State $score^{\star} \, , \, split^{\star} \, , \, feature^{\star}\gets score \, , \,  x_{i_{w^{\star}},j} \, , \,  j$
    \EndIf
\EndFor
\State {\bf Return} $split^{\star},feature^{\star}$
\end{algorithmic}
\end{algorithm}

\begin{algorithm}
\caption{GetEntropiesPMQRT}\label{algo:GetEntropiesPMQRT}
\begin{algorithmic}
    \State {\bf Input}: List of values in the leaf $\textbf y$
    \State Initialize empty min-max heaps $\mathcal H_m$ for $m\in \{1,\dots,M+1\}$ and insert $y_1$ in $\mathcal H_1$.
    \State $\ell_{\mathcal E} \gets  $ [0 \, , \, get\_entropy$((\mathcal H_m)_{m\in[M+1]}, 1)]$  {\color{blue}$/\star$\text{ cf. Algorithm \ref{alg:get_entropyPMQRT}}}   
    \For{$s=2$ to $n_k$}
        \State $\tilde y \leftarrow y_{i_s}$
        \For{$m=1$ to $M$}
            \If{$\sum_{\ell=1}^m |\mathcal H_{\ell}|=\ceil{\tau_m s}+1$}
                \If{$\mathcal H_m$ is not empty and $\tilde y \leq \max(\mathcal H_m)$}
                    \State Remove the maximum $A_{j,k,s,m}$ of $\mathcal H_m$.
                    \State Insert $\tilde y$ in $\mathcal H_m$.
                    \State $\tilde y \leftarrow A_{j,k,s,m}$.
                \EndIf
            \Else{
                \State $\widehat m$, $\widehat a_{j,k,s,m}$ $\gets $ get\_min\_heaps\_plus$(m+1)$ {\color{blue}$/\star$\text{ cf. Algorithm \ref{alg:get_min_heaps_plus}}}                    \If{($\widehat a_{j,k,s,m}$ is $None$) or $(\tilde y \leq \widehat a_{j,k,s,m}$) }
                    \State Insert $\tilde y$ in $\mathcal H_m$.
                    \If{$\widehat a_{j,k,s,m}$ is $None$}
                        \State $\widetilde y \gets None$
                        \State break
                    \Else
                        \State Remove the min $\widehat a_{j,k,s,m}$ of $\mathcal H_{\widehat m}$
                        \State $\widetilde y \gets \widehat a_{j,k,s,t}$ 
                    \EndIf
                \Else
                    \State Remove the min $\widehat a_{j,k,s,m}$ of $\mathcal H_{\widehat m}$
                    \State Insert $\widehat a_{j,k,s,m}$ in $\mathcal H_t$
                \EndIf
            }
            \EndIf
        \EndFor
        \If{$\tilde y$ is not $None$}
            \State Insert $\tilde y$ in $\mathcal H_{M+1}$.
        \EndIf
    \State $\ell_{\mathcal E}.append\big($get\_entropy$((\mathcal H_m)_{m\in[M+1]}, s)\big)$  {\color{blue}$/\star$\text{ cf. Algorithm \ref{alg:get_entropyPMQRT}}}   
    \EndFor
\end{algorithmic}
\end{algorithm}

\begin{algorithm}
\caption{Get max value among heaps below a given index.}\label{alg:get_max_heaps_minus}
\begin{algorithmic}
\Procedure{get\_max\_heaps\_minus}{idx\_heap}
  \State max\_val = None
  \While{(idx\_heap $\geq 0$) and (max\_val is None)}
        \State max\_val $\gets$ $\max(\mathcal H_{\text{idx\_heap}})$
        \If{max\_val is None}
            \State idx\_heap $\gets$ idx\_heap $- 1$
        \EndIf
    \EndWhile
    \State {\bf return} idx\_heap, max\_val
\EndProcedure
\end{algorithmic}
\end{algorithm}

\begin{algorithm}
\caption{Get min value among heaps above a given index.}\label{alg:get_min_heaps_plus}
\begin{algorithmic}
\Procedure{get\_min\_heaps\_plus}{idx\_heap}
  \State min\_val = None
  \While{(idx\_heap $<M+1$) and (min\_val is None)}
        \State min\_val $\gets$ $\min(\mathcal H_{\text{idx\_heap}})$
        \If{min\_val is None}
            \State idx\_heap $\gets$ idx\_heap $+ 1$
        \EndIf
    \EndWhile        
    \State {\bf return} idx\_heap, min\_val
\EndProcedure
\end{algorithmic}
\end{algorithm}

\begin{algorithm}
\caption{Get current entropy from the min-max heaps.}\label{alg:get_entropyPMQRT}
\begin{algorithmic}
\Procedure{get\_entropy}{$(\mathcal H_m)_{m\in[M+1]}$, $n$}
  \State entropy = 0
  \State heap2sums $\gets (\mathcal H_m.sum()$ for $m\in [M+1])$
  \State total\_sum $\gets$ sum(heap2sums)
  \State cum\_sum $\gets$ cumsum(heap2sums)
  \For{$m=1$ to $M$}
     \State $\_, $ max\_heap\_m = get\_max\_heaps\_minus$(m)$ {\color{blue}$/\star$\text{ cf. Algorithm \ref{alg:get_max_heaps_minus}}}   
    \State entropy += max\_heap\_m . $(\ceil{n\tau_m}/n-\tau_m)$
    \State entropy += tot\_sum . $\tau_m / n$
    \State entropy -= cum\_sum[m] $/ n$
\EndFor
    \State {\bf return} entropy
\EndProcedure
\end{algorithmic}
\end{algorithm}

\clearpage
\subsection{Computations of the entropies for all splits with the CRPS}
\label{apdx:crps_induction}

Theorem 1 ensures that one can compute the information gain at a given node for the possible split of the data points in that node by:
\begin{itemize}
\item[1.] computing the insertion ranks $(r_s)_{s\in[n]}$ before running the iterative scheme. This can be done in time complexity $\mathcal O(n\log n)$ using a weight-balanced binary tree. 
\item[2.] updating along the iterative process a Fenwick tree $\mathcal F$ that will allow us at iteration $s$ to query $S^{(s)}_{r_s}, S^{(s)}$ and $W^{(s)}$ in $\mathcal O(\log n)$ time.
\end{itemize}

Algorithm~\ref{alg:crps_tree} describes the procedure to compute the information gain induced by all the possible splits at a given node.

\begin{algorithm}
\caption{CRPS-RT: Returns the feature index and the threshold value giving the split of the data with the largest information gain.}\label{alg:crps_tree}
\begin{algorithmic}[1]
 \State {\bf Input}: Data in a given node of the leaf $\{(\textbf x_i,y_i)\}_{i=1}^n$
\State $score^{\star}, split^{\star}, feature^{\star}\gets + \infty, None, None$
\For{$j=1$ to $d$}
    \State Let $x_{i_1 ,j} \geq \dots \geq  x_{i_n,j}$ be the decreasing sorted $(x_{i,j} )_{i\in [n]}$ and set $\textbf y:=\big( y_{i_1},\dots,y_{i_{n}}\big).$
    \State $\sigma \gets \mathrm{ArgSort}(\textbf y) $\;
    \State  $\textbf r \gets \mathrm{GetRanks}(\textbf y)$ \quad {\color{blue}$/\star$\text{ cf. Lemma 1}} 
    \State $\ell_{\mathcal E}^{\downarrow} \gets $  GetEntropiesCRPS($\textbf y,\sigma,\textbf r)$ \quad {\color{blue}$/\star$\text{ cf. Algorithm~\ref{algo:GetEntropiesCRPS}}} \vspace{0.2cm}
    \State Let $x_{i_1 ,k} \leq \dots \leq  x_{i_{n},k}$ be the increasing sorted $(x_{i,k} )_{i \in [n]}$ and set $\textbf y:=\big( y_{i_1},\dots,y_{i_{n}}\big).$
    \State $\sigma \gets \mathrm{ArgSort}(\textbf y) $\;
    \State  $\textbf r \gets \mathrm{GetRanks}(\textbf y)$  \quad {\color{blue}$/\star$\text{ cf. Lemma 1}} 
    \State $\ell_{\mathcal E}^{\uparrow} \gets $ GetEntropiesCRPS($\big( y_{i_1},\dots,y_{i_{n}}\big),\sigma,\textbf r)$ \quad {\color{blue}$/\star$\text{ cf. Algorithm~\ref{algo:GetEntropiesCRPS}}} \vspace{0.2cm}
    
    \State $w^{\star} , score \gets \text{ArgMin}\big(w\times\ell_{\mathcal E}^{\uparrow}[w]+(n-w)\times\text{Flip}(\ell_{\mathcal E}^{\downarrow})[w], \quad w \in \{0,\dots, n\}\big)$
    \If{$score < score^{\star}$}
        \State $score^{\star} \, , \, split^{\star} \, , \, feature^{\star}\gets score \, , \,  x_{i_{w^{\star}},j} \, , \,  j$
    \EndIf
\EndFor
\State {\bf Return} $split^{\star},feature^{\star}$
\end{algorithmic}
\end{algorithm}

\begin{algorithm}
\caption{GetEntropiesCRPS}\label{algo:GetEntropiesCRPS}
\begin{algorithmic}[1]
    \State {\bf Input}: List of values in the leaf $\textbf y$, Permutation sorting the list $\sigma$, ranks $\textbf r$
    \State\tikzmark{start1} $S,W , h \gets 0$ 
    \State $\mathcal F \gets$ Empty Fenwick Trees of size $n$    
    \State $\ell_{\mathcal E} \gets [0]$ \tikzmark{end1}\vspace{0.13cm}
    \For{$s=1$ to $n$} 
            
        \State\tikzmark{start2} $ W \gets W  +r_{s}y_{s} + S-\mathcal F\mathrm{.get}(1,\sigma^{-1}(s))$
        \State $S \gets S+y_{s}$
        \State $h \gets h + (s-1)(2r_s-3-(s-1))y_s+2W-2S-2(s-1)\mathcal F\mathrm{.get}(1,\sigma^{-1}(s))$
        \State $\ell_{\mathcal E}.append(\, h / s^3\,)$ \quad {\color{blue}$/\star$\text{ or if using LOO and }$s\neq 1:\, h/ [s(s-1)^2]$}\hspace{0cm}\tikzmark{end2}\vspace{0.2cm} 
        \State \tikzmark{start3}$\mathcal F\mathrm{.add}(\sigma^{-1}(s),y_{s})$ \quad {\color{blue}$/\star$\text{ cf. Algorithm~\ref{alg:add_tree}}}
        \hspace{-7.45cm}\tikzmark{end3}         \vspace{0.1cm}
    \EndFor
    \State {\bf Return} $\ell_{\mathcal E}$
\end{algorithmic}
\Textbox[7cm]{start1}{end1}{Initialization}
\Textbox[3cm]{start2}{end2}{Compute entropy}
\Textbox[10cm]{start3}{end3}{Update tree}
\end{algorithm}

\begin{algorithm}
\caption{Add for Fenwick Tree.}\label{alg:add_tree}
\begin{algorithmic}
    \State {\bf Input}: position $idx$, value $z$, length $n$
     \While{$idx\leq n$}
        \State $ self.tree[idx-1] \gets  self.tree[idx-1]  + z$
        \State $idx = idx +  (idx$ 
 $\&^1-idx)$
     \EndWhile
     \State {\color{blue}/$\star$ $^1$The operator $\&$ compares the binary representations of the two numbers, bit by bit, returning a new integer where each bit is set to 1 only if both bits in the same position are 1.}
\end{algorithmic}
\end{algorithm}

\section{Proofs}

\subsection{Proof of Theorem 1}
The CRPS is defined by
\[\mathrm{CRPS}(F,y) = \int_{-\infty}^{+\infty} \big( F(s) - \mathds 1_{s\geq y}\big)^2 ds.\]
The associated entropy is 
\[ \min_F \mathds E[ \mathrm{CRPS}(F,Y)].\]
Considering its empirical counterpart, the minimizer is the empirical CDF.
\[\hat F:s\mapsto \frac1n \sum_{i=1}^n \mathds 1_{s\geq y_i}.\]
We deduce that the empirical entropy is given by
\allowdisplaybreaks
\begingroup
\begin{align*}
&\frac1n \sum_{i=1}^n\mathrm{CRPS}(\widehat F,y_i)\\
& = \frac1n \sum_{i=1}^n \int_{-\infty}^{+\infty} \big( \hat F(s) - \mathds 1_{s\geq y_i}\big)^2 ds\\
&=   \frac1n \sum_{i=1}^n \int_{-\infty}^{+\infty} \big( \frac1n \sum_{j=1}^n \mathds 1_{s\geq y_j} - \mathds 1_{s\geq y_i}\big)^2 ds\\
&=   \frac{1}{n^3} \sum_{i,j,k} \int_{-\infty}^{+\infty} \big( \mathds 1_{s\geq y_j} - \mathds 1_{s\geq y_i}\big)\big( \mathds 1_{s\geq y_k} - \mathds 1_{s\geq y_i}\big)  ds\\
&=   \frac{2}{n^3} \sum_{i<j<k} \int_{-\infty}^{+\infty} \big( \mathds 1_{s\geq y_{(j)}} - \mathds 1_{s\geq y_{(i)}}\big)\big( \mathds 1_{s\geq y_{(k)}} - \mathds 1_{s\geq y_{(i)}}\big)  ds\\
&\qquad + \frac{2}{n^3} \sum_{j<i<k} \int_{-\infty}^{+\infty} \big( \mathds 1_{s\geq y_{(j)}} - \mathds 1_{s\geq y_{(i)}}\big)\big( \mathds 1_{s\geq y_{(k)}} - \mathds 1_{s\geq y_{(i)}}\big)  ds\\
&\qquad + \frac{2}{n^3} \sum_{j<k<i} \int_{-\infty}^{+\infty} \big( \mathds 1_{s\geq y_{(j)}} - \mathds 1_{s\geq y_{(i)}}\big)\big( \mathds 1_{s\geq y_{(k)}} - \mathds 1_{s\geq y_{(i)}}\big)  ds\\
&\qquad + \frac{2}{n^3} \sum_{j<i} \int_{-\infty}^{+\infty} \big( \mathds 1_{s\geq y_{(j)}} - \mathds 1_{s\geq y_{(i)}}\big)^2  ds\\
&=   \frac{2}{n^3} \sum_{i<j<k}  (y_{(j)}-y_{(i)})  \quad + 0\quad  + \frac{2}{n^3} \sum_{j<k<i}  (y_{(i)}-y_{(k)})+ \frac{2}{n^3} \sum_{j<i} (y_{(i)}-y_{(j)}) \\ 
&=   \frac{2}{n^3} \sum_{i=1}^{n-2} \sum_{j=i+1}^{n-1}  (n-j)(y_{(j)}-y_{(i)}) - \frac{2}{n^3} \sum_{j=1}^{n-2} \sum_{k=j+1}^{n-1}  \sum_{i=k+1}^{n}(y_{(k)}-y_{(i)})+\frac{2}{n^3} \sum_{i<j} (y_{(j)}-y_{(i)})\\ 
&=   \frac{2}{n^3} \sum_{i=1}^{n-1} \sum_{j=i+1}^{n}  (n-j+1)(y_{(j)}-y_{(i)}) - \frac{2}{n^3} \sum_{j=1}^{n-2} \sum_{k=j+1}^{n-1}  \sum_{i=k+1}^{n}(y_{(k)}-y_{(i)})\\ 
&=  - \frac{2}{n^3} \sum_{i=1}^{n-1} [n(n-i) -\frac{n(n-1)}{2}+\frac{i(i-1)}{2}] y_{(i)}  + \frac{2}{n^3}\sum_{j=2}^{n}(n-j+1)(j-1)y_{(j)}\\
&\qquad -\frac{2}{n^3} \sum_{i=3}^{n} \sum_{k=2}^{i-1} \sum_{j=1}^{k-1}  (y_{(k)}-y_{(i)})\\ 
&= -  \frac{2}{n^3} \sum_{i=1}^{n-1} \frac12(n-i)(n-i+1) y_{(i)}  + \frac{2}{n^3}\sum_{j=2}^{n}(n-j+1)(j-1)y_{(j)}-\frac{2}{n^3} \sum_{i=3}^{n} \sum_{k=2}^{i-1} (k-1)  (y_{(k)}-y_{(i)})\\ 
&=  - \frac{1}{n^3} \sum_{i=1}^{n-1} (n-i)(n-i+1) y_{(i)}  + \frac{2}{n^3}\sum_{j=2}^{n}(n-j+1)(j-1)y_{(j)}\\
&\qquad +\frac{2}{n^3} \sum_{i=3}^{n} \frac{(i-2)(i-1)}{2} y_{(i)} - \frac{2}{n^3} \sum_{k=2}^{n-1} (n-k)(k-1)y_{(k)} \\ 
&= -  \frac{1}{n^3} \sum_{i=1}^{n-1} (n-i)(n-i+1) y_{(i)}  +\frac{1}{n^3} \sum_{i=3}^{n} (i-2)(i-1) y_{(i)} +\frac{2}{n^3} \sum_{k=2}^{n} (k-1)y_{(k)}\\
&= -  \frac{1}{n^3} \sum_{i=1}^{n} (n-i)(n-i+1) y_{(i)}  +\frac{1}{n^3} \sum_{i=3}^{n} (i-2)(i-1) y_{(i)} +\frac{2}{n^3} \sum_{k=1}^{n} (k-1)y_{(k)}\\
&= \frac{1}{n^3} \sum_{i=1}^{n} \left[(i-2)(i-1)+2(i-1)-(n-i)(n-i+1) \right] y_{(i)}  \\
&= \frac{1}{n^3} \sum_{i=1}^{n} \left[i(i-1)-(n-i)(n-i+1) \right] y_{(i)}  \\
&=   \frac{1}{n^3} \sum_{i=1}^{n} (i-1)i (y_{(i)}-y_{(n-i+1)}) .
  \end{align*}
\endgroup
We denote by $\textbf y^{(n)}$ the vector of size $n$ corresponding to the $n$ first entries of the vector of observations $\textbf y$ (where we consider without loss of generality that elements in $\textbf y$ have been ordered considering the permutation sorting in increasing order $\big( x_{i,k}\big)_{i: \textbf x_i \in R_j}$ for some given node $k$ and some feature $j$). From the previous computations we have 
\[\frac1n \sum_{i=1}^n\mathrm{CRPS}(\widehat F,y_i) = \frac{1}{n^3} \left(h^{(n)}_{\uparrow} -h^{(n)}_{\downarrow} \right),\]
where \begin{align*}
h^{(n)}_{\uparrow}&:=\sum_{i=1}^{n} (i-1)iy^{(n)}_{(i)}\quad \text{and} \quad
h^{(n)}_{\downarrow}:=\sum_{i=1}^{n} (i-1)i y^{(n)}_{(n-i+1)}.
\end{align*}
In the following, we derive iterative formula for both $h^{(n)}_{\uparrow}$ and $h^{(n)}_{\downarrow}$.

\paragraph{Iterative formula for $h^{(n)}_{\uparrow}$}.

\allowdisplaybreaks
\begingroup
\begin{align*}
h^{(n+1)}_{\uparrow}-h^{(n)} _{\uparrow}&=\sum_{i=1}^{n+1} (i-1)iy^{(n+1)}_{(i)} -\sum_{i=1}^{n} (i-1)iy^{(n)}_{(i)} \\
&=\sum_{i=1}^{n+1} (i-1)i y^{(n+1)}_{(i)} -\sum_{i=1}^{r_{n+1}-1} (i-1)iy^{(n+1)}_{(i)} -\sum_{i=r_{n+1}}^{n} (i-1)i y^{(n+1)}_{(i+1)} \\
&=\sum_{i=1}^{n+1} (i-1)iy^{(n+1)}_{(i)} -\sum_{i=1}^{r_{n+1}-1} (i-1)i y^{(n+1)}_{(i)} -\sum_{i=r_{n+1}+1}^{n+1} (i-2)(i-1) y^{(n+1)}_{(i)} \\
&=(r_{n+1}-1) r_{n+1} y^{(n+1)}_{(r_{n+1})} + 2\sum_{i=r_{n+1}+1}^{n+1} (i-1) y^{(n+1)}_{(i)}\\
&=[(r_{n+1}-1) r_{n+1} - 2(r_{n+1}-1)] y^{(n+1)}_{(r_{n+1})} + 2 W^{(n+1)}\\
&-2\sum_{i=1}^{r_{n+1}-1} i y^{(n+1)}_{(i)}-2S^{(n+1)}+2\sum_{i=1}^{r_{n+1}-1}  y^{(n+1)}_{(i)}\\
&=[(r_{n+1}-1) r_{n+1} - 2(r_{n+1}-1)] y^{(n+1)}_{(r_{n+1})} + 2 W^{(n+1)}\\
&-2\sum_{i=1}^{r_{n+1}-1} i y^{(n+1)}_{(i)}-2S^{(n+1)}+2S^{(n+1)}_{r_{n+1}-1}.
\end{align*}
\endgroup

\paragraph{Iterative formula for $h^{(n)}_{\downarrow}$}.

\allowdisplaybreaks
\begingroup
\begin{align*}
h^{(n+1)}_{\downarrow}-h^{(n)}_{\downarrow} &=\sum_{i=1}^{n+1} (i-1)i y^{(n+1)}_{(n-i+2)} -\sum_{i=1}^{n} (i-1)i y^{(n)}_{(n-i+1)} \\
&=\sum_{i=1}^{n+1} (n-i+1)(n-i+2) y^{(n+1)}_{(i)} -\sum_{i=1}^{n} (n-i)(n-i+1) y^{(n)}_{(i)} \\
&=\sum_{i=1}^{n+1}  (n-i+1)(n-i+2)y^{(n+1)}_{(i)} -\sum_{i=1}^{r_{n+1}-1} (n-i)(n-i+1) y^{(n+1)}_{(i)}\\
&\quad -\sum_{i=r_{n+1}}^{n} (n-i)(n-i+1) y^{(n+1)}_{(i+1)} \\
&=\sum_{i=1}^{n+1}  (n-i+1)(n-i+2) y^{(n+1)}_{(i)} -\sum_{i=1}^{r_{n+1}-1} (n-i)(n-i+1) y^{(n+1)}_{(i)}\\
&\quad -\sum_{i=r_{n+1}+1}^{n+1} (n-i+1)(n-i+2)y^{(n+1)}_{(i)} \\
&= 2\sum_{i=1}^{r_{n+1}-1}   (n-i+1)y^{(n+1)}_{(i)} + (n-r_{n+1}+1) (n-r_{n+1}+2) y^{(n+1)}_{(r_{n+1})} \\
&= 2(n+1) S^{(n+1)}_{r_{n+1}-1}-2 \sum_{i=1}^{r_{n+1}-1} i y^{(n+1)}_{(i)} + (n-r_{n+1}+1) (n-r_{n+1}+2) y^{(n+1)}_{(r_{n+1})} .
\end{align*}
\endgroup

\paragraph{Conclusion.}
Using the previous computations, we obtain that 
\begin{align*}
h^{(n+1)}_{\uparrow}-h^{(n+1)}_{\downarrow}
&=h^{(n)}_{\uparrow}-h^{(n)}_{\downarrow}
 + n\big[2r_{n+1} -3-n\big] y^{(n+1)}_{(r_{n+1})} + 2 W^{(n+1)}-2S^{(n+1)}-2nS^{(n+1)}_{r_{n+1}-1}.
\end{align*}
Noticing that $S^{(n+1)}_{r_{n+1}-1}=S^{(n+1)}_{r_{n+1}}-y_{n+1}$ concludes the proof.

\subsection{Proof of Lemma 2}
\label{proof:lemma:fenwick_tree}

\paragraph{Querying $S^{(s)}$ in $\mathcal O(\log n)$ time.}

Let us recall that by construction, $\mathcal F$ is at iteration $s$ a Fenwick tree for the sequence $\big( \mathrm 1_{\sigma(i)\leq s} y_{\sigma(i)}\big)_{i\in[n]}$. One can thus easily see that getting the sum of all elements in $\mathcal F$ gives $S^{(s)}=\sum_{i=1}^s y^{(s)}_{(i)}=\sum_{i=1}^s y_i $. Indeed, up to ordering of the elements, the list $\big( \mathrm 1_{\sigma(i)\leq s} y_{\sigma(i)}\big)_{i\in[n]}$ is equal to the one obtained by applying the permutation $\sigma^{-1}$ namely to $\big( \mathrm 1_{i\leq s} y_{i}\big)_{i\in[n]}$ whose sum is equal to $S^{(s)}$. Hence, querying the sum of all elements of $\mathcal F$ at iteration $s$ gives $S^{(s)}$ and this operation requires $ \mathcal O(\log n)$ time since $\mathcal F$ is a Fenwick tree.

\paragraph{Querying $S^{(s)}_{r_s-1}$ in $\mathcal O(\log n)$ time.}

Let us now prove that querying the sum of all elements in $\mathcal F$ from position $1$ to $\sigma^{-1}(s)$ at iteration $s$ is equal to $S^{(s)}_{r_s}$. This will prove that $S^{(s)}_{r_s}$ can be obtained in $\mathcal O(\log n)$ time since $\mathcal F$ is a Fenwick tree. One can then get $S^{(s)}_{r_s-1}$ using that $y^{(s)}_{(r_s)}=y_s$ and thus that
\[S^{(s)}_{r_s-1}=S^{(s)}_{r_s} -y_s.\]
Indeed, $S^{(s)}_{r_s-1}=\sum_{i=1}^{r_s-1}y^{(s)}_{(i)}$ where $r_s$ is the position of insertion of $y_s$ in the non-decreasing sorted version of the list $\mathbf y^{(s-1)}$.

\bigskip
Since $y^{(s)}_{(r_s)}=y_s$, the list $(y^{(s)}_{(1)}, \dots,y^{(s)}_{(r_s)})$ exactly corresponds to all the elements in $(y_{1}, \dots, y_{s})$ smaller than $y_s$ i.e. to the list of all $y_k$ for $k\leq s$ such that $\sigma^{-1}(k)\leq \sigma^{-1}(s)$. We deduce that
\begin{equation}\label{eq:lemma2} S^{(s)}_{r_s}=\sum_{i=1}^{r_s} y^{(s)}_{(i)} = \sum_{i=1}^s \mathrm 1_{\sigma^{-1}(i)\leq \sigma^{-1}(s)} y_i=\sum_{j\in \{ \sigma^{-1}(k) \mid k\in[s]\}} \mathrm 1_{j\leq \sigma^{-1}(s)} y_{\sigma(j)}=\sum_{j=1}^{\sigma^{-1}(s)} \mathrm 1_{\sigma(j)\leq s} y_{\sigma(j)},\end{equation}
where in the penultimate equality we made the change of variable $i=\sigma(j)$ and where in the last equality we used that 
\[ \{ \sigma^{-1}(k) \mid k\in[s], \sigma^{-1}(k)\leq \sigma^{-1}(s)\}= \{ j\mid j\in[\sigma^{-1}(s)], \sigma(j)\leq s\}.\]
Since $\mathcal F$ is at iteration $s$ a Fenwick tree for the sequence $\big( \mathrm 1_{\sigma(i)\leq s} y_{\sigma(i)}\big)_{i\in[n]}$, we deduce from Eq.\eqref{eq:lemma2} that querying the sum of elements from position 1 to $\sigma^{-1}(s)$ in $\mathcal F$ gives $S^{(s)}_{r_s}$.

As a remark, one can notice that a direct consequence of this result is that querying the sum of all elements from position 1 to $\sigma^{-1}(s+1)$ at iteration $s$ from $\mathcal F$ gives $S^{(s)}_{r_{s+1}-1}$.

\paragraph{Getting $W^{(s+1)}$ in constant time given $W^{(s)}$, $S^{(s)}$ and $S^{(s+1)}_{r_{s+1}-1}$.}

The total rank-weighted sum $W^{(s)}$ can be efficiently tracked along the iterative process using the formula
\begin{equation}\label{eq:key_update_CRPS_apdx}W^{(s+1)} = W^{(s)} +r_{s+1}y_{s+1} +\sum_{i=r_{s+1}}^s y^{(s)}_{(i)}=W^{(s)}  +r_{s+1}y_{s+1} +S^{(s)}-S^{(s)}_{r_{s+1}-1},\end{equation}
where 
\[S^{(s)}_{r_{s+1}-1} = S^{(s+1)}_{r_{s+1}-1},\]
since $r_{s+1}$ is the position of insertion of $y_{s+1}$ in the list $( y^{(s)}_{(1)}, \dots, y^{(s)}_{(s)})$.

\section{Unbiased estimate of information gains}

\label{apdx:LOO}

\subsection{Reminder: Risk estimation for the quadratic loss}

Consider a dataset $\mathcal D_n={(x_i,y_i)}_{i=1}^n$ with squared loss
$\ell(\hat y,y)=(y-\hat y)^2$. Let $\hat f_{1:n}$ denote the estimator trained on the full dataset and $\hat f^{-i}_{1:n}$ the estimator trained on the sample with observation $i$ removed.

\paragraph{Mallows $C_p$ and optimism correction for the quadratic loss.}

A natural estimator of the prediction risk of $\hat f_{1:n}$ is the empirical risk
\[
\widehat R_{1:n}(\hat f_{1:n})=\frac{1}{n}\sum_{i=1}^n (y_i-\hat f_{1:n}(x_i))^2 .
\]
However, this quantity is optimistically biased because the same data are used both to fit the model and evaluate its performance. In expectation,
\[
\mathbb{E}[\widehat R_{1:n}(\hat f_{1:n})] \le \mathbb{E}[R(\hat f_{1:n})],
\]

where $R(\hat f_{1:n})=\mathbb{E}_{(X,Y)}[(Y-\hat f_{1:n}(X))^2]$ denotes the population risk.

A general way to correct this bias is to introduce the optimism term
\[
\mathrm{pen}_{n} := \mathbb{E}\left[ R(\hat f_{1:n})-\widehat R_{1:n}(\hat f_{1:n}) \right].
\]
This quantity measures the expected gap between the true prediction risk and the training error. If an unbiased estimate $ \widehat{\mathrm{pen}}_{1:n}$ of $\mathrm{pen}_n$ is available, one can construct an unbiased estimator of the prediction risk:
\[
\widehat R_{1:n}(\hat f_{1:n}) + \widehat{\mathrm{pen}}_{1:n} .
\]
Mallows $C_p$ provides such an estimator in the case of squared loss under the classical homoscedastic regression model. For linear estimators with effective degrees of freedom $d$ and noise variance $\sigma^2$, the optimism admits the simple expression
\[
\mathrm{pen}_n = \frac{2\sigma^2}{n} d .
\]
This leads to the well-known $C_p$ criterion
\[
C_p = \widehat R_{1:n}(\hat f_{1:n}) + \frac{2\sigma^2}{n} d .
\]

\paragraph{Leave-One-Out (LOO) estimate.}

The leave-one-out estimate of the prediction risk is defined as
\[
\widehat R^{\text{LOO}}_{1:n} := \frac{1}{n}\sum_{i=1}^{n} (y_i - \hat f^{-i}_{1:n}(x_i))^2 .
\]
Each term evaluates the prediction error of a model trained without observation $i$ on that left-out observation. $\widehat R^{\text{LOO}}$ is an unbiased estimate of the expected population risk of the learning algorithm trained on $n-1$ observations,
\[\mathbb{E}\big[ R(\hat f_{1:n-1}) \big],
\]
where $R(\hat f_{1:n-1})=\mathbb{E}_{(X,Y)}[(Y-\hat f_{1:n-1}(X))^2]$ denotes the population risk of the estimator $\hat f_{1:n-1}$ trained with $n-1$ observations.

Importantly, this means that LOO does not provide an unbiased estimate of the risk of $\hat f_{1:n}$ (the estimator trained on the full dataset). The two risks are typically very close when $n$ is large, but they correspond to different training sample sizes.

\subsection{Unbiased estimate of information gain in regression trees.}

Several authors have recognized that the standard CART algorithm can produce biased estimates of the information gain because splits are evaluated on the same data used to estimate node parameters. In particular, S. Athey et al.\footnote{Footnote 6: Wager, S., \& Athey, S. (2018). Estimation and inference of heterogeneous treatment effects using random forests. Journal of the American Statistical Association, 113(523), 1228-1242.} already noted that one should use strategies, such as Mallows-style corrections, to adjust for the additional sampling variance in small leaves. Implementing such a correction in practice requires efficiently estimating the corresponding optimism term for each possible split and feature. 

In the case of the quadratic loss, the standard Mallows $C_p$ correction terms does not induce additional computational complexity since the estimate of $\mathrm{pen}_{n}$ given by $\widehat{\mathrm{pen}}_{1:n}:=\frac{2\widehat \sigma^2}{n}d$ where $\widehat \sigma^2$ is the standard unbiased estimate of the noise variance. Similarly, the LOO estimate of the information gain for the quadratic loss does not cause any computational overhead.

The choice between Mallows-style correction and LOO should be guided by their respective theoretical properties and computational costs. Suppose we are evaluating a node split for a leaf containing $n$ observations. We denote by $\hat f_{k:j}^{\mathrm{ERM}}$ the empirical risk minimizer for the loss $\ell$ (typically the CRPS of the WIS) computing from the data points $k$ to $j$ (with $1\leq k< j\leq n$). With the notations introduced in the main paper, we have $ \widehat R_{k:j}(\hat f^{\mathrm{ERM}}_{k:j})=H_{\ell}(\mathbf y_{k:j})$. Rather than relying on a standard empirical estimate of information gain, we can enhance the decision procedure using one of the following approaches:
\begin{itemize}
\item Mallows-style correction provides an unbiased estimate of the expected information gain for the model trained on the same number of data points as the one the would be used at inference time. Namely, for a candidate split leading to the children nodes with data $(y_1,\dots,y_k)$ and $(y_{k+1},\dots,y_n)$, the standard information gain
\begin{align*}\widehat { \mathrm{IG}}_k&:=\widehat R_{1:n}(\hat f_{1:n}^{\mathrm{ERM}})-\left(\widehat R_{1:k}(\hat f_{1:k}^{\mathrm{ERM}}) + \widehat R_{k:n}(\hat f_{k:n}^{\mathrm{ERM}})\right)\\
&= H_{\ell}(\mathbf y_{1:n}) - (H_{\ell}(\mathbf y_{1:k}) + H_{\ell}(\mathbf y_{k:n}))
\end{align*}
can be corrected by adding the appropriate optimism terms, namely one adds $\widehat{\mathrm{pen}}_{1:n}$ to $H_{\ell}(\mathbf y_{1:n})$, $\widehat{\mathrm{pen}}_{1:k}$ to $H_{\ell}(\mathbf y_{1:k})$ and  $\widehat{\mathrm{pen}}_{k:n}$ to $H_{\ell}(\mathbf y_{k:n})$. It holds
\begin{equation}  \mathbb E\left[\widehat{\mathrm{IG}}_k+\widehat{\mathrm{pen}}_{1:n}-\left(\widehat{\mathrm{pen}}_{1:k}+\widehat{\mathrm{pen}}_{k:n}\right) \right]=\mathbb E\left[R(\hat f_{n}^{\mathrm{ERM}}) - \left(R(\hat f_{k}^{\mathrm{ERM}}) + R(\hat f_{n-k}^{\mathrm{ERM}})\right)\right].\label{eq:target_IG}\end{equation}

\item LOO estimates of the information gains yield unbiased estimates of the risk for models trained one less data point in each child leaf. Stated otherwise, the expected value of the LOO estimate of the information gain is given by:
\[ \mathbb E\left[ R(\hat f_{n-1}^{\mathrm{ERM}}) - \left( R(\hat f_{k-1}^{\mathrm{ERM}}) +  R(\hat f_{n-k-1}^{\mathrm{ERM}})\right)\right]\]
Consequently, LOO provides a biased estimate for the quantity we are targeting, namely the one given in Eq.\eqref{eq:target_IG}.

However, this bias tends to penalize splits that create very small leaves, because removing a single observation from a small leaf has a larger effect on the LOO error. As a result, LOO-based selection often favors splits that produce larger, more balanced leaves, leading to more robust trees and potentially better generalization performance. Our numerical experiments, presented in the main paper, indicate that the results obtained using either LOO or Mallows corrections are very consistent.
\end{itemize}

In the rest of this section, we show that LOO estimates of the information gains are available for both the sum of pinball losses and the CRPS loss without additional computational cost. We also show that a Mallows-style correction can be derived for the CRPS loss at essentially no extra cost. Deriving a similar Mallows-type correction for the sum of pinball losses appears to be more challenging. We therefore do not provide such a result here, and to the best of our knowledge we are not aware of any existing work that derives such a correction.

\subsubsection{LOO for PMQRT}
\label{apdx:LOO_PMQRT}

We propose using a leave-one-out method to obtain an unbiased estimate of the information gain used to build the PQRT or PMQRT. In the following, we show that such LOO estimate of the entropy associated with a pinball loss $\ell_{\tau}$ with data $\textbf y=(y_1,\dots,y_n)$ can be computed efficiently.

Let us denote by $\mathcal H_n$ the entropy of the sequence $\textbf y$ and by $\mathcal H_{n,LOO}$ the leave-one-out estimate of the entropy of $\textbf y$ for the pinball loss $\ell_{\tau}$, namely
\[\mathcal H_n = \frac{1}{n}\sum_{i=1}^n \ell_{\tau}(y_i, \hat q_{\tau}),\quad \mathcal H_{n,LOO} = \frac{1}{n}\sum_{i=1}^n \ell_{\tau}(y_i, \hat q_{\tau}^{-i}),\]
where $\hat q_{\tau}$ (resp. $\hat q_{\tau}^{-i}$) is the empirical quantile of order $\tau$ of the sequence $\textbf y$ (resp. $\textbf y^{-i}:=(y_1,\dots, y_{i-1},y_{i+1},\dots, y_n)$).

Let us consider a permutation $\sigma$ sorting the sequence $\textbf y$ by increasing order. We further denote $r_i:=\sigma^{-1}(i)$ which corresponds to the position in the sorted list of the element $y_i$. Defining $r^-:=\ceil{\tau (n-1)}$ and $r^*:=\ceil{n\tau}$, we can distinguish two possible cases:
\begin{itemize}
\item If $r^-=r^*$:
    \begin{itemize}
    \item If $r_i>r^-$:  $\ell_{\tau}(y_i, \hat q^{-i}_{\tau})=\ell_{\tau}(y_i, \hat q_{\tau}) = \tau (y_i - y_{(r^*)}).$
    \item Else: $\ell_{\tau}(y_i, \hat q^{-i}_{\tau})=(1-\tau)(y_{(r^*+1)}-y_i).$
    \end{itemize}
    Hence:
    \begin{align*}\mathcal H_{n,LOO} &= \mathcal H_n - \frac{(1-\tau)}{n} \sum_{i \mid r_i\leq r^*} (y_{(r^*)} -y_i) + \frac{(1-\tau)}{n} \sum_{i \mid r_i\leq r^*} (y_{(r^*+1)} -y_i) \\
    &= \mathcal H_n + \frac{(1-\tau) r^*}{n} (y_{(r^*+1)} - y_{(r^*)} ).
    \end{align*}
\item Else (i.e. if $r^-+1=r^*$):
    \begin{itemize}
    \item If $r_i>r^-$:  $\ell_{\tau}(y_i, \hat q^{-i}_{\tau})=\ell_{\tau}(y_i, y_{(r^*-1)}) = \tau (y_i - y_{(r^*-1)}).$
    \item Else: $\ell_{\tau}(y_i, \hat q^{-i}_{\tau})=(1-\tau)(y_{(r^*)}-y_i).$
    \end{itemize}
     Hence:
    \begin{align*}\mathcal H_{n,LOO} &= \mathcal H_n - \frac{\tau}{n} \sum_{i \mid r_i\leq r^*} (y_i-y_{(r^*)} ) + \frac{\tau}{n} \sum_{i \mid r_i\leq r^*}  (y_i-y_{(r^*-1)} )\\
    &= \mathcal H_n + \frac{\tau (n-r^*+1)}{n} (y_{(r^*)} - y_{(r^*-1)} ). \end{align*}
\end{itemize}

All quantities $y_{(r^*-1)}, y_{(r^*)}$ and $y_{(r^*+1)}$ can be queried in $\mathcal O(\log(n))$ time using the min-max heap structures used to train PMQRT (or PQRT). Overall, a node split using or not these LOO estimates of entropies is performed in $\mathcal O(d n \log(n))$ time where $d$ is the number of features and $n$ is the number of data points in the leaf. Therefore, no additional cost is induced by using LOO.

\begin{figure*}[!ht]
    \centering
    \begin{subfigure}[t]{0.5\textwidth}
        \centering
        \includegraphics[height=2.2in]{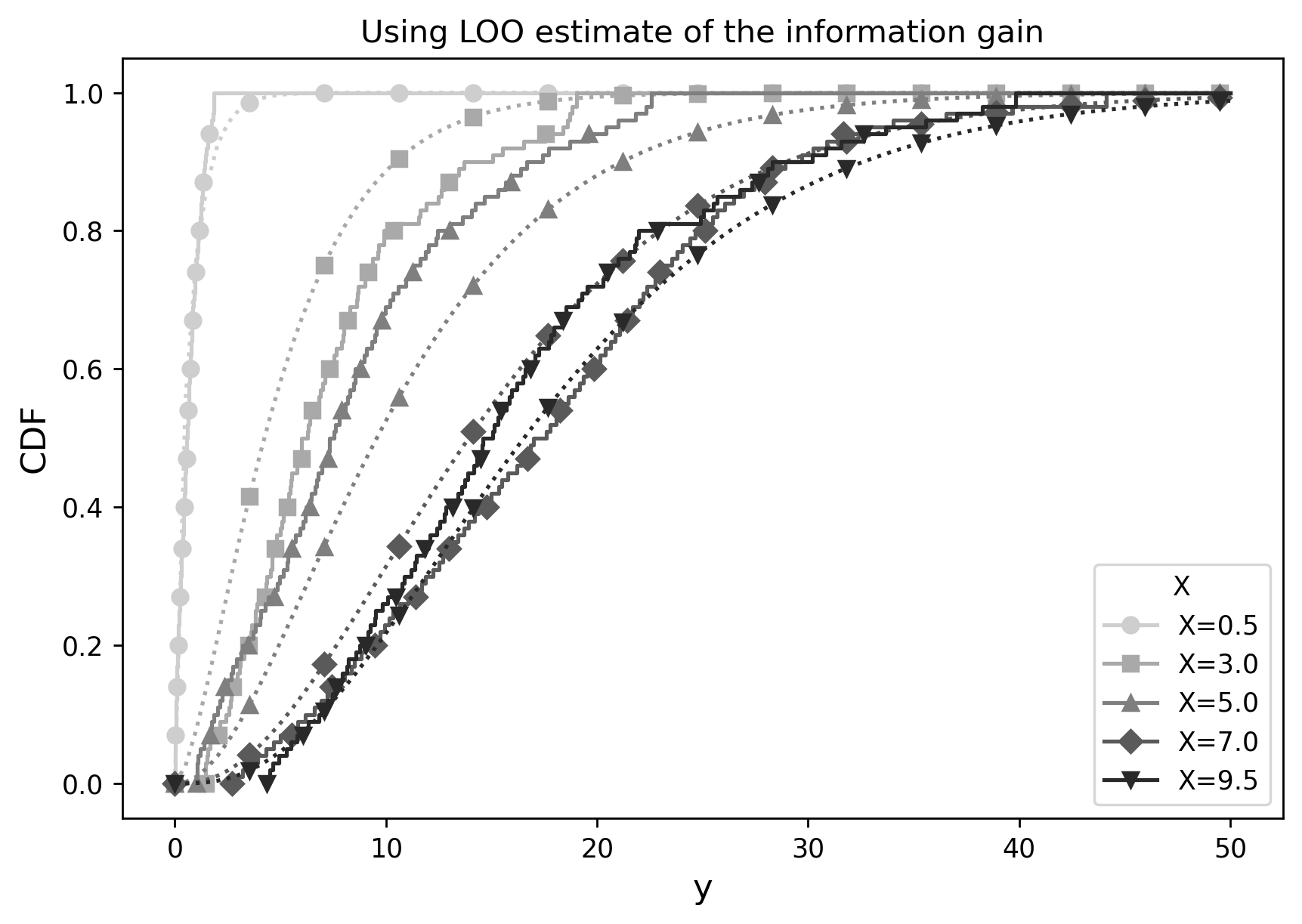}
        \caption{With LOO.}
    \end{subfigure}%
    ~ 
    \begin{subfigure}[t]{0.5\textwidth}
        \centering
        \includegraphics[height=2.2in]{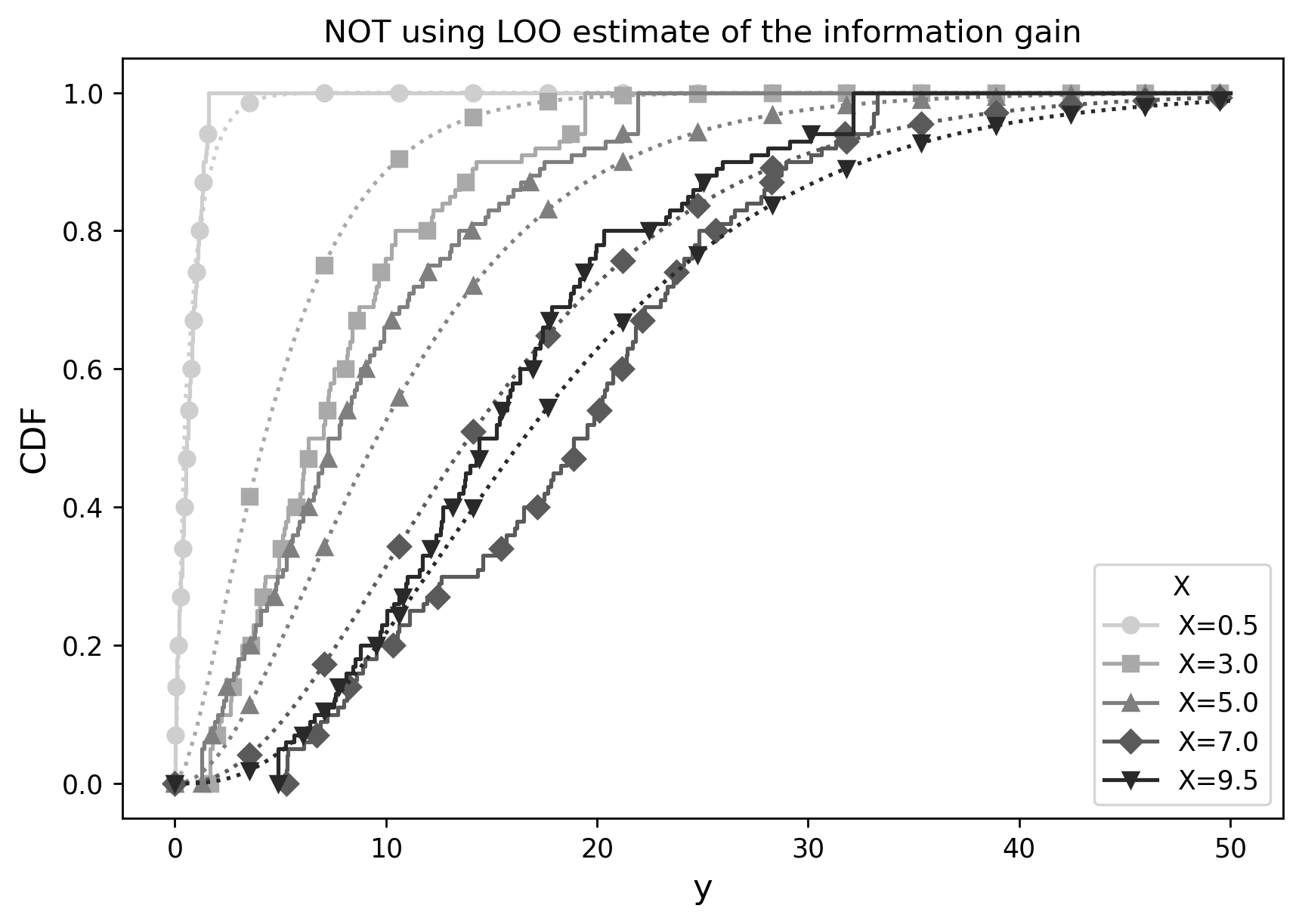}
        \caption{Without LOO.}
    \end{subfigure}
    \caption{Estimation of the conditional CDF of y given x with PMQRF with $100$ on a sample of size $n=600$ simulated from the distribution in Eq.(6) of the main manuscript. The true conditional CDFs (smooth dashed lines) are compared to the quantile estimates obtained from PMQRF (dots). The maximum depth was set 10 and the  minimum number of samples required to split an internal node (min\_sample\_split) was set to 10.}
    \label{fig:simu_LOO_pmqrt}
\end{figure*}

Figure~\ref{fig:simu_LOO_pmqrt} illustrates on a numerical example the difference on the estimated conditional quantiles when learning PMQRTs using or not without.

\subsubsection{LOO for CRPS-RT}
It is noteworthy that using leave-one-out to obtain an unbiased estimate of the information gain at each node for determining the best split is very simple. In fact, the computations presented below show that the leave-one-out estimate of the information gain can be efficiently derived by appropriately scaling the standard empirical entropies.

Denoting by $\hat F^{-i}$ the empirical CDF obtained from the sequence $(y_1,\dots,y_{i-1},y_{i+1},\dots,y_n)$, we have
\allowdisplaybreaks
\begingroup
\begin{align*}
&\frac1n \sum_{i=1}^n\mathrm{CRPS}(\widehat F^{-i},y_i)\\
& = \frac1n \sum_{i=1}^n \int_{-\infty}^{+\infty} \big( \hat F^{-i}(s) - \mathds 1_{s\geq y_i}\big)^2 ds\\
&=   \frac1n \sum_{i=1}^n \int_{-\infty}^{+\infty} \big( \frac{1}{n-1} \sum_{j\neq i} \mathds 1_{s\geq y_j} - \mathds 1_{s\geq y_i}\big)^2 ds\\
&=   \frac{1}{n(n-1)^2} \sum_{i=1}^n \sum_{j\neq i} \sum_{k\neq i}\int_{-\infty}^{+\infty} \big( \mathds 1_{s\geq y_j} - \mathds 1_{s\geq y_i}\big)\big( \mathds 1_{s\geq y_k} - \mathds 1_{s\geq y_i}\big)  ds\\
&=   \frac{1}{n(n-1)^2} \sum_{i=1}^n \sum_{j=1}^n  \sum_{k=1}^n\int_{-\infty}^{+\infty} \big( \mathds 1_{s\geq y_j} - \mathds 1_{s\geq y_i}\big)\big( \mathds 1_{s\geq y_k} - \mathds 1_{s\geq y_i}\big)  ds\\
&= \frac{n^2}{(n-1)^2} \left(\frac1n \sum_{i=1}^n\mathrm{CRPS}(\widehat F,y_i) \right).
  \end{align*}
\endgroup

\subsubsection{Optimism correction for CRPS-RT}

We derive here the analogue of Mallows' $C_p$ for the CRPS risk when the predictive distribution is the empirical distribution function
\[
\widehat F(s)=\frac1n\sum_{i=1}^n \mathds 1_{\{Y_i\le s\}},
\]
constructed from an i.i.d.\ sample $\mathcal D_n=(Y_1,\dots,Y_n)$ with common distribution function $F$. 

For a fixed predictive CDF $\widehat F$, the empirical CRPS criterion is
\begin{align*}
\widehat { R}_{1:n}(\widehat F)=H_{\mathrm{CRPS}}(\mathbf y) &= \frac1n \sum_{i=1}^n \int_{-\infty}^{+\infty} \left(\widehat F(s) - \mathds 1_{s\geq y_i}\right)^2 ds \\
&= \int_{-\infty}^{+\infty} \left( \widehat F(s)^2 -2 \widehat F(s) \frac{1}{n}\sum_{i=1}^n\mathds 1_{s\geq y_i} + \frac{1}{n}\sum_{i=1}^n\mathds 1_{s\geq y_i}\right) ds \\
&= \int_{-\infty}^{+\infty} \left(  \widehat F(s)-\widehat F(s)^2 \right) ds .
\end{align*}
The corresponding out-of-sample (prediction) risk is
\begin{align*}
 R(\widehat F) &= \mathbb E_{Y,\mathcal D_n}\left[\int_{-\infty}^{+\infty} \left(\widehat F(s) - \mathds 1_{s\geq Y}\right)^2 ds\right] \\
&= \int_{-\infty}^{+\infty} \mathbb E_{\mathcal D_n}\left[ \widehat F(s)^2  - 2 \widehat F(s) \mathbb E_{Y}[\mathds 1_{s\geq Y} ] +  \mathbb E_{Y}[\mathds 1_{s\geq Y}^2 ] \right]ds\\
&= \int_{-\infty}^{+\infty} \mathbb E_{\mathcal D_n}\left[ \widehat F(s)^2  - 2 \widehat F(s) F(s) +  F(s) \right]ds\\
&= \int_{-\infty}^{+\infty}\left( \mathbb E_{\mathcal D_n}\left[ \widehat F(s)^2 \right] - 2  F(s)^2 +  F(s) \right) ds.
\end{align*}
where $Y$ is an independent future draw from $F$. We used that $\mathbb E[\mathds 1_{\{Y\le s\}}]=F(s)$ and $\mathbb E[\mathds 1_{\{Y\le s\}}^2]=F(s)$. Therefore the expected optimism is
\begin{align*}
\mathrm{pen}_n&=\mathbb E_{\mathcal D_n}\left[ R(\widehat F) - \widehat { R}_{1:n}(\widehat F)  \right] \\
&= \int_{-\infty}^{+\infty}\left(2\mathbb E_{\mathcal D_n}\left[\widehat F(s)^2 \right]-2F(s)^2 \right)ds = \int_{-\infty}^{+\infty} 2 \mathrm{Var}_{\mathcal D_n}(\widehat F(s))ds\\
&= \int_{-\infty}^{+\infty} 2 \mathrm{Var}_{\mathcal D_n}(\frac1n \sum_{i=1}^n \mathds 1_{s\geq y_i}) ds\\
&= \int_{-\infty}^{+\infty} \frac2n \mathrm{Var}_{\mathcal D_n}( \mathds 1_{s\geq y_1}) ds\\
&= \int_{-\infty}^{+\infty} \frac2n F(s)(1-F(s))ds.
\end{align*}

This suggests the plug-in estimator of $\mathrm{pen}_n$ given by:
\[\widehat{\mathrm{pen}}_{1:n}: =\int_{-\infty}^{+\infty} \frac{2}{n-1} \widehat F(s)(1-\widehat F(s))ds =\frac{2}{n-1} \widehat R _{1:n}(\widehat F)=\frac{2}{n-1} H_{\mathrm{CRPS}}(\mathbf y). \]

Normalization by $n-1$ ensures that the plug-in estimator is exactly unbiased. Indeed, for every $s\in\mathbb R$,
\[
\mathbb E_{\mathcal D_n}[\widehat F(s)]=F(s)
\qquad\text{and}\qquad
\mathrm{Var}_{\mathcal D_n}(\widehat F(s))=\frac{1}{n}F(s)\bigl(1-F(s)\bigr),
\]
so that
\begin{align*}
\mathbb E_{\mathcal D_n}\!\left[\widehat F(s)\bigl(1-\widehat F(s)\bigr)\right]
&= \mathbb E_{\mathcal D_n}[\widehat F(s)]-\mathbb E_{\mathcal D_n}[\widehat F(s)^2] \\
&= F(s)-\Bigl(\mathrm{Var}_{\mathcal D_n}(\widehat F(s))+\mathbb E_{\mathcal D_n}[\widehat F(s)]^2\Bigr) \\
&= F(s)-\frac1nF(s)(1-F(s))-F(s)^2 \\
&= \left(1-\frac1n\right)F(s)\bigl(1-F(s)\bigr).
\end{align*}
Integrating over $s$ yields
\[
\mathbb E_{\mathcal D_n}\!\left[H_{\mathrm{CRPS}}(\mathbf y)\right]
=
\left(1-\frac1n\right)\int_{-\infty}^{+\infty}F(s)\bigl(1-F(s)\bigr)\,ds
=
\frac{n-1}{n}\int_{-\infty}^{+\infty}F(s)\bigl(1-F(s)\bigr)\,ds.
\]
Therefore,
\begin{align*}
\mathbb E_{\mathcal D_n}\!\left[\frac{2}{n-1}H_{\mathrm{CRPS}}(\mathbf y)\right]
&=
\frac{2}{n-1}\cdot \frac{n-1}{n}\int_{-\infty}^{+\infty}F(s)\bigl(1-F(s)\bigr)\,ds \\
&=
\frac{2}{n}\int_{-\infty}^{+\infty}F(s)\bigl(1-F(s)\bigr)\,ds
=
\mathrm{pen}_n,
\end{align*}
which proves that the normalization by $n-1$ yields an exactly unbiased estimator of the optimism term.

Hence the counterpart of Mallows $C_p$ for the CRPS loss is given by:
\begin{align*}&H_{\mathrm{CRPS}}(\mathbf y) + \frac{2}{n-1} H_{\mathrm{CRPS}}(\mathbf y) = \frac{n+1}{n-1}H_{\mathrm{CRPS}}(\mathbf y).
\end{align*}
In other words, CRPS-RT admits a simple closed-form optimism correction, so that model selection can be based on a direct multiplicative rescaling of the in-sample CRPS criterion.

In the regression tree setting, this correction is applied locally at each candidate split: the distribution function $F$ should therefore be understood as the conditional cumulative distribution function of the response $Y$ given that the covariate vector $X$ belongs to the node (or child node) defined by the sequence of previous splits together with the split currently under consideration, so that the resulting Mallows-style correction provides an unbiased estimate of the corresponding CRPS information gain.

\section{Datasets}

Table~\ref{tab:datasetsUCI} lists the datasets used in our experiments, along with their corresponding URLs.

\begin{table}[!ht]
\centering
\begin{tabular}{|c|c|c|c|}\hline
{\bf Dataset} & $n$  & $d$ & URL \\\hline
Abalone & 4177 & 8 & \href{https://archive.ics.uci.edu/dataset/1/abalone}{abalone} \\\hline
GPU & $241600 $ & $ 14$ & \href{https://archive.ics.uci.edu/dataset/440/sgemm+gpu+kernel+performance}{sgemm+gpu+kernel+performance}\\ \hline
Turbine &  $36733$ & $12$  & \href{https://archive.ics.uci.edu/dataset/551/gas+turbine+co+and+nox+emission+data+set}{gas+turbine+co+and+nox+emission+data+set} \\ \hline
Cycle & $9568$ & $4$ & \href{https://archive.ics.uci.edu/dataset/294/combined+cycle+power+plant}{combined+cycle+power+plant}  \\ \hline
WineRed & $4898$ & $11$ & \href{https://archive.ics.uci.edu/dataset/186/wine+quality}{wine-quality/winequality-red.csv}\\\hline
WineWhite & $1599$ & $11$ & \href{https://archive.ics.uci.edu/dataset/186/wine+quality}{wine-quality/winequality-white.csv} \\ \hline
\end{tabular}
\caption{Meta-data for the datasets. $n$ refers to the total
number of data-points from which we create $300$ versions by independently drawing $1000$ data points
randomly (as described in the beginning of Section 3.3 of the main paper). $d$ refers to the feature dimension.}
\label{tab:datasetsUCI}
\end{table}

\end{document}